\documentclass[10pt,twocolumn,letterpaper]{article}

\usepackage{cvpr}              

\definecolor{clova}{rgb}{0.24, 0.63, 0.33}

\newcommand{\hboh}[1]{\textcolor{black}{#1}}

\definecolor{CuGray}{gray}{0.5}
\newcommand{\gray}[1]{\textcolor{CuGray}{#1}}

\usepackage[accsupp]{axessibility}

\usepackage[utf8]{inputenc} %
\usepackage[T1]{fontenc}    %
\usepackage[pagebackref,breaklinks,colorlinks,allcolors=cvprblue]{hyperref}
\usepackage{url}            %
\usepackage{booktabs}       %
\usepackage{amsfonts}       %
\usepackage{nicefrac}       %
\usepackage{microtype}      %
\usepackage[dvipsnames]{xcolor}         %
\usepackage{notoccite}
\usepackage{titletoc}

\usepackage{colortbl}
\usepackage{graphicx,overpic}
\usepackage[export]{adjustbox}
\usepackage{amsmath, amssymb}
\usepackage{multirow}
\usepackage[super]{nth}
\usepackage{comment}
\usepackage{nicematrix}

\usepackage{enumitem}
\usepackage{caption}

\usepackage{titlesec}

\titleformat{\section}
{\normalfont\large\bfseries}{\thesection.~}{0pt}{}{}
\titleformat{\subsection}
{\normalfont\large\bfseries}{\thesubsection.~}{0pt}{}{}
\titleformat{\subsubsection}
{\normalfont\normalsize\bfseries}{\thesubsubsection.~}{0pt}{}{}
\titleformat{\paragraph}[runin]
{\normalfont\normalsize\bfseries}{\theparagraph\vspace{1em}}{}{}

\titlespacing*{\section}{0pt}{2.0ex plus .5ex minus .5ex}{7pt plus 2px minus 2px}
\titlespacing*{\subsection}{0pt}{1.0ex plus .5ex minus .5ex}{5pt plus 2px minus 2px}
\titlespacing*{\subsubsection}{0pt}{.5ex plus .5ex minus .5ex}{3pt plus 1pt minus 1pt}
\titlespacing*{\paragraph}{0em}{.5ex plus .5ex minus .3ex}{1em}

\usepackage{tikz}
\usepackage{pgfplots}
\usepackage{pgfplotstable}
\pgfplotsset{compat=1.13}
\usetikzlibrary{pgfplots.groupplots}
\usetikzlibrary{calc,arrows}

\newcommand\pmnum[1]{\small$\pm$#1}

\newcommand*\fdvgg{FD\(_{\text{VGG}}\)}
\newcommand*\fdpann{FD\(_{\text{PANNs}}\)}
\newcommand*\fdpasst{FD\(_{\text{PaSST}}\)}
\newcommand*\ispann{IS}

\newcommand*\klpann{KL\(_{\text{PANNs}}\)}
\newcommand*\klpasst{KL\(_{\text{PaSST}}\)}

\usepackage{tabularx,stackengine,collcell}
\let\endminwd\relax
\newcolumntype{L}[1]{>{\collectcell\xminwd l{#1}}l<{\endminwd\endcollectcell}}
\newcolumntype{C}[1]{>{\collectcell\xminwd c{#1}}c<{\endminwd\endcollectcell}}
\newcolumntype{R}[1]{>{\collectcell\xminwd r{#1}}r<{\endminwd\endcollectcell}}
\def\minwd#1#2#3\endminwd{\stackengine{0pt}{#3}{\rule{#2}{0pt}}{O}{#1}{F}{F}{L}}
\newcommand\xminwd[1]{\minwd#1}

\hypersetup{
	colorlinks,
	linkcolor={red!80!black},
	citecolor={cvprblue},
	urlcolor={cvprblue},
    pdftitle={PAVAS: Physics-Aware Video-to-Audio Synthesis}
}

\definecolor{defaultColor}{RGB}{230, 244, 252}

\usepackage{bbding}
\usepackage{pifont}
\usepackage{wasysym}
\usepackage{amssymb}

\expandafter\def\expandafter\normalsize\expandafter{%
    \normalsize%
    \setlength{\abovedisplayskip}{6pt}
    \setlength{\belowdisplayskip}{6pt}
    \setlength{\abovedisplayshortskip}{6pt}
    \setlength{\belowdisplayshortskip}{6pt}
}

\definecolor{cvprblue}{rgb}{0.21,0.49,0.74}
\usepackage[pagebackref,breaklinks,colorlinks,allcolors=cvprblue]{hyperref}

\def\paperID{6748} 
\def\confName{CVPR}
\def\confYear{2026}

\title{PAVAS: Physics-Aware Video-to-Audio Synthesis}

\def\authorBlock{
    Oh Hyun-Bin${}^{1\dagger}$\enspace
    Yuhta Takida${}^{2}$\enspace
    Toshimitsu Uesaka${}^{2}$\enspace
    Tae-Hyun Oh${}^{4}$\enspace
    Yuki Mitsufuji${}^{2,3}$\vspace{3mm} \\
   \small{${}^{1}$POSTECH\quad${}^{2}$Sony AI\quad${}^{3}$Sony Group Corporation\quad${}^{4}$KAIST} \vspace{2mm}\\ 
}

\begin{document}
\author{\authorBlock}

\maketitle

\def\thefootnote{\textdagger}\footnotetext{Work done during an internship at Sony AI.}
\def\thefootnote{\arabic{footnote}}

\begin{abstract}
Recent advances in Video-to-Audio (V2A) generation have achieved impressive perceptual quality and temporal synchronization, yet most models remain appearance-driven, capturing visual-acoustic correlations without considering the physical factors that shape real-world sounds.
We present Physics-Aware Video-to-Audio Synthesis (PAVAS), a method that incorporates physical reasoning into a latent diffusion-based V2A generation through the Physics-Driven Audio Adapter (Phy-Adapter).
The adapter receives object-level physical parameters estimated by the Physical Parameter Estimator (PPE), which uses a Vision-Language Model (VLM) to infer the moving-object mass and a segmentation-based dynamic 3D reconstruction module to recover its motion trajectory for velocity computation.
These physical cues enable the model to synthesize sounds that reflect underlying physical factors.
To assess physical realism, we curate VGG-Impact, a benchmark focusing on object–object interactions, and introduce Audio-Physics Correlation Coefficient (APCC), an evaluation metric that measures consistency between physical and auditory attributes.
Comprehensive experiments show that PAVAS produces physically plausible and perceptually coherent audio, outperforming existing V2A models in both quantitative and qualitative evaluations.
\gray{Visit https://physics-aware-video-to-audio-synthesis.github.io.}
\end{abstract}
    
\section{Introduction}
\label{sec:intro}

Humans effortlessly infer physical properties of the world from both what they see and what they hear~\cite{opoku2021visual},
and visual information influences humans' auditory perception due to the audiovisual integration property~\cite{fujisaki2014audiovisual} and prediction mechanism~\cite{schroger2015attention}.
For example, when a hammer strikes metal or a ball bounces on a floor, we intuitively expect the resulting sound to reflect underlying physical factors such as object velocity, mass, and material.

\begin{figure}
    \centering
    \includegraphics[width=1.0\linewidth]{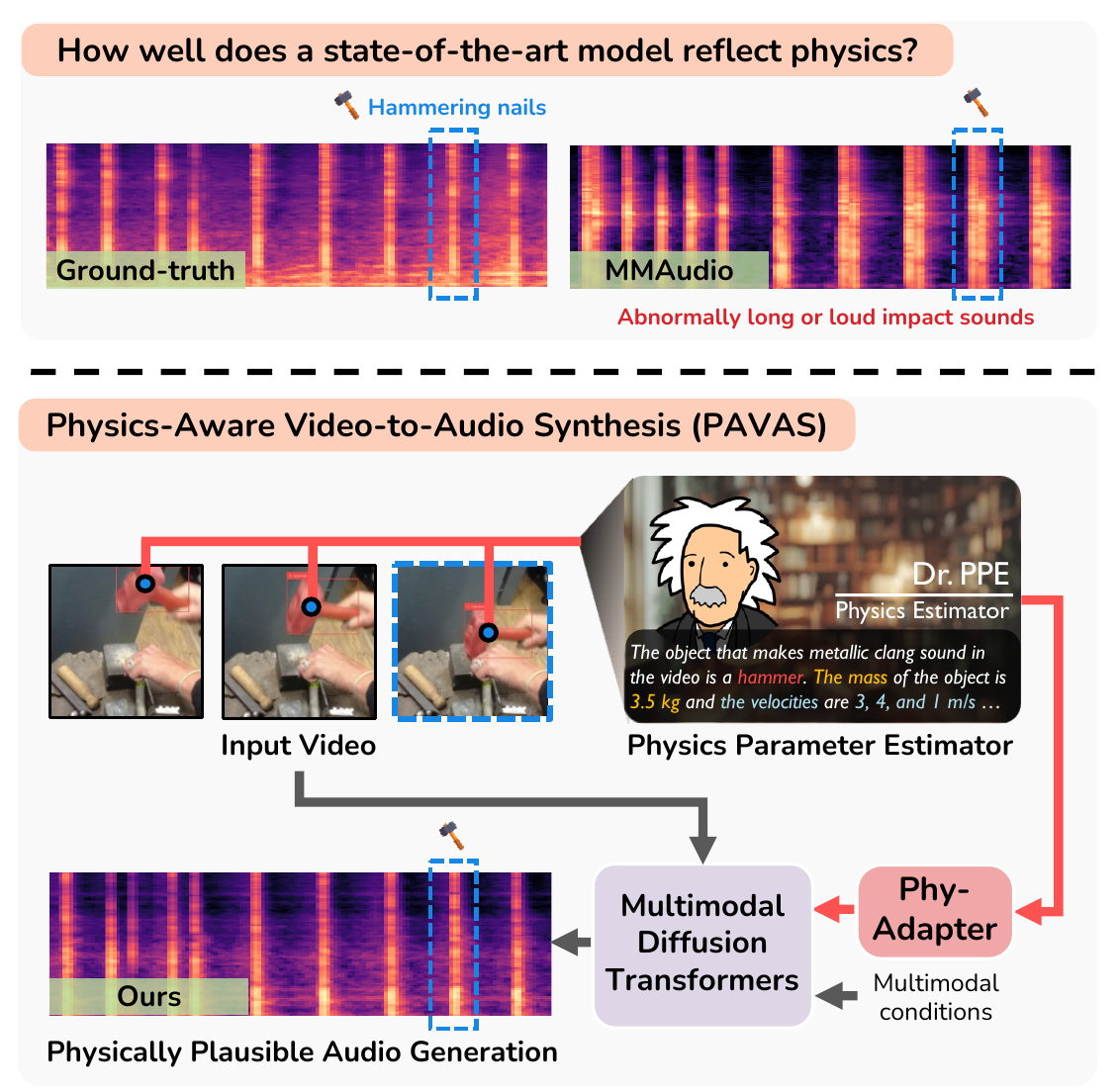}
    \caption{\textbf{Physics-Aware Video-to-Audio Synthesis (PAVAS)}. \textbf{[Top]} Current V2A models often generate physically inconsistent audio. \textbf{[Bottom]} We estimate physics values (object-level mass and velocity) from an input video using Physics Parameter Estimator, which are explicitly integrated into a latent diffusion-based model using Phy-Adapter to generate a physically plausible audio.}
    \label{fig:teaser}
\end{figure}

Agnostic about this, recent Video-to-Audio (V2A) generation models, including autoregressive-\cite{sheffer2023hear,liu2024tell,viertola2025temporally}, GAN-\cite{iashin2021taming}, and diffusion-based~\cite{luo2024diff,xing2024seeing,zhang2024foleycrafter,wang2025frieren,pham2024mdsgen,jeong2025read,wang2024v2a,ton2025taro} approaches, learn to generate audio that aligns temporally and semantically with video content and 
have achieved impressive perceptual quality and synchronization, especially with the emergence of latent diffusion frameworks~\cite{rombach2022high,esser2024scaling,labs2025flux1kontextflowmatching}. 
Yet, they remain appearance-driven: a model may correctly associate a hammering motion with a ``metallic clang,'' but it may fail to modulate the loudness or spectral sharpness according to the strength and dynamics of the impact, producing physically implausible audio. (see Fig.~\ref{fig:teaser}-[Top]).

We refer to this discrepancy as a lack of \emph{physical grounding}—implicit modeling between visual dynamics and acoustic behavior.
We define physics-aware audio generation as the synthesis of sounds that not only align with visual events but also produce sounds whose acoustic properties vary consistently with measurable physical values, such as object mass and velocities in a video. 
Our work bridges this gap by explicitly incorporating object-level physical cues into a V2A generation, allowing the model to synthesize sounds that are \emph{perceptually coherent} and \emph{physically consistent}.

To this end, we propose Physics-Aware Video-to-Audio Synthesis (PAVAS), which integrates explicit physical reasoning into a latent diffusion-based generation process (see Fig.~\ref{fig:teaser}-[Bottom]).
PAVAS includes two key modules: (i) a Physics Parameter Estimator (PPE) that extracts object-level physical quantities from videos, and (ii) a Physics-Driven Audio Adapter (Phy-Adapter), a modulation module that injects the estimated physical parameters into the diffusion model to guide sound synthesis. The PPE consists of a Mass Estimator, which leverages a Vision-Language Model (VLM)~\cite{bai2025qwen2} to infer object mass from visual and semantic context, and a Velocity Estimator, which combines a text-grounded segmentation model~\cite{ren2024grounded} and a dynamic 3D reconstruction model~\cite{wang2025continuous} to recover object-level motion and estimate velocity.  
Surprisingly, we find that our estimators achieve physics value estimation performance comparable to, or even surpassing, that of specialized expert models~\cite{nerf2physics,chen2024spatialvlm}, in both the mass and velocity estimations.
Through these components, PAVAS allows the generation process to reflect the object dynamics of real-world interactions, capturing how object motion and mass influence the resulting sound.

Furthermore, we identify the need for an evaluation protocol that explicitly measures physical realism in video-to-audio generation.
Existing benchmarks such as VGGSound~\cite{chen2020vggsound} mainly assess perceptual or semantic alignment, and fail to capture whether generated sounds are consistent with the underlying physical dynamics.
To address this, we curate a VGG-Impact benchmark, focusing on object--object interaction events (\eg, collisions, impacts, or bouncing), where physical cues play a critical role.
We also present the Audio-Physics Correlation Coefficient (APCC), which quantifies the correlation between estimated physical values (\ie, kinetic energy) and generated audio's attributes (\ie, spectral energy), providing an interpretable measure of physical grounding beyond existing perceptual metrics.

Extensive experiments on VGGSound~\cite{chen2020vggsound} and VGG-Impact demonstrate that our approach substantially improves physical plausibility while maintaining high perceptual quality.
Our model outperforms existing V2A baselines~\cite{jeong2025read,xing2024seeing,viertola2025temporally,liu2024tell,wang2025frieren,zhang2024foleycrafter,wang2024v2a,cheng2025mmaudio,ton2025taro} on both VGGSound and human evaluations, generating high-quality sounds that are semantically and temporally aligned with video content.
Further analysis on VGG-Impact shows that, unlike prior appearance-driven models that exhibit weak or inconsistent correlations on 
APCC,
PAVAS captures physically meaningful relationships between object motion and sound, achieving the closest APCC to ground-truth data.
Our main contributions are summarized as follows:
\begin{itemize}
\item We introduce PAVAS, a Physics-Aware Video-to-Audio Synthesis pipeline that injects object-level physical parameters—estimated by our reliable physics estimators—into a latent diffusion model via the proposed Phy-Adapter.
\item We curate VGG-Impact, a novel benchmark for object–object interaction sounds, and propose APCC, a metric designed to evaluate the physical consistency between visual dynamics and a generated audio.
\item We conduct comprehensive analyses of existing V2A models on both VGGSound and VGG-Impact, revealing that prior work often produce physically inconsistent sounds, whereas our method achieves more physically consistent and perceptually realistic audio generation.
\end{itemize}

\section{Related Work}
\label{sec:formatting}

\paragraph{Video-to-audio synthesis}
Video-to-audio (V2A) generation aims to synthesize realistic sounds that are temporally synchronized and semantically aligned with video content. Recent approaches~\cite{iashin2021taming,sheffer2023hear,luo2024diff,xing2024seeing,zhang2024foleycrafter,wang2025frieren,pham2024mdsgen,cheng2025mmaudio,ton2025taro,karchkhadze2025stereofoley,zhao2025foleyspace} have significantly advanced this task using increasingly expressive generative models.
Autoregressive methods~\cite{iashin2021taming,sheffer2023hear} model sequential dependencies effectively, while diffusion-based~\cite{ho2020denoising, song2020denoising} approaches further improve perceptual quality and synchronization~\cite{ho2020denoising,song2020denoising,luo2024diff,xing2024seeing,zhang2024foleycrafter,wang2025frieren,pham2024mdsgen,ton2025taro}. More recently, MMAudio~\cite{cheng2025mmaudio} leverages large-scale text-audio data together with limited video data, establishing a strong foundation for modern V2A synthesis.

Despite these advances, existing audio-visual cross generation methods remain appearance-driven, relying on visual--acoustic correlations~\cite{senocak2023sound,senocak2025toward,sung2024sound2vision,sung2023sound,sung2024sound2vision,sung2025soundbrush,chae2025perceptually} without modeling the underlying physical dynamics of real-world interactions.
Auxiliary conditioning signals such as onsets, motion energy, or mel cues~\cite{jeong2025read,ton2025taro,cheng2025mmaudio} help synchronization, but these cues often produce smoothed features that fail to capture fine-grained visual dynamics.
We instead move beyond perceptual alignment toward physics-aware sound generation, incorporating explicit object-level physical parameters into a V2A generation process.

\paragraph{Physics parameter estimation from video}
Inferring physical properties from visual data has long been a core goal in visual physics reasoning. 
Early work explore predicting object dynamics and material attributes such as mass, friction, or elasticity from visual cues or simulated interactions~\cite{fragkiadaki2016learning,wu2015galileo,yildirim2016interpreting,bell2015material}.
In this work, we focus on mass and velocity, two quantities that directly determine the magnitude and temporal evolution of physical events in videos. 
While appearance already encodes material-related cues, mass and velocity govern how objects move, collide, and produce sound, enabling physically aware audio generation.

Recent advances in visual mass estimation, such as NeRF2Physics~\cite{nerf2physics}, combine neural radiance fields~\cite{mildenhall2020nerf,jun2024factorized,kim2024fprf,jun2022hdr} with vision-language features~\cite{radford2021learning} to infer physical properties from multi-view images. However, these methods require static, calibrated views and cannot operate on dynamic videos. In contrast, we leverage physics knowledge embedded in Vision-Language Models (VLMs)~\cite{bai2025qwen2,hyeon2024vlm,ye2024beaf,sung2024avhbench,ye2025retouchllm} to estimate object mass from
a monocular video, 
achieving comparable performance while maintaining open-world generalization.
Estimating object velocity has been studied less extensively.
I-MOVE~\cite{schwan2020move} segments independently moving regions to infer per-object velocity, while most other methods rely on active sensors such as radar or LiDAR~\cite{guo2022doppler} are limited to vehicle-centric domains~\cite{kampelmuhler2018camera,sormoli2024optical}. 
Building on recent progress in open-vocabulary segmentation~\cite{kirillov2023segment,liu2023groundingdino} and dynamic 3D reconstruction~\cite{wang2025continuous}, our method estimates reliable physical parameters from unconstrained videos, enabling physics-aware conditioning for video-to-audio synthesis.

\paragraph{Injecting physics cues in video-to-audio generation}
Recent efforts in Video-to-Audio (V2A) generation have begun to incorporate physical-condition signals to enhance realism. Su~\etal\cite{su2023physicsdiffusion} introduce a diffusion-based model conditioned on physics priors to synthesize impact sounds from silent videos; however, their approach is restricted to generating a drumstick sound and therefore does not generalize to diverse object--object interactions.
Saad~\etal\cite{saad2025materialacoustic} generate acoustic profiles for indoor scenes by controlling material parameters, and SonifyAR~\cite{su2024sonifyar} proposes an AR sound-authoring pipeline that uses contextual cues such as surface material and interaction type to produce spatialized effects. 
While these works incorporate contextual or material cues, they are largely designed for indoor or AR-focused settings and do not explicitly model physical factors such as object-level mass or motion dynamics.

\section{Physics-Aware Video-to-Audio Synthesis}
\label{sec:method}
We propose \textbf{Physics-Aware Video-to-Audio Synthesis (PAVAS)}, a novel approach for generating audio that is consistent not only with the visual context but also with the underlying physical dynamics present in a video. 
PAVAS is based on a latent diffusion architecture and integrates explicit object-level physical reasoning to ensure the generated sounds are physically plausible.

We first describe the diffusion-based backbone and its multimodal conditioning interface (Sec.~\ref{sec:preliminary}). 
We then provide an overview of the complete pipeline (Sec.~\ref{sec:overview}), followed by detailed descriptions of the two key modules: 
the \textit{Physics Parameter Estimator} (PPE; Sec.~\ref{sec:extractor}), which extracts object-level mass and velocity from a video, 
and the \textit{Physics-Driven Audio Adapter} (Phy-Adapter; Sec.~\ref{sec:phyda}), which incorporates these physical cues into the diffusion model.

\subsection{Preliminary}
\label{sec:preliminary}
We formulate video-to-audio generation as a latent modeling problem. First, we prepare a variational autoencoder (VAE)~\cite{kingma2013auto} pretrained on the mel-spectrogram domain, as well as a pretrained vocoder~\cite{lee2022bigvgan} that converts mel-spectrograms into waveform audio signals. 
By combining the VAE decoder with the vocoder, compressed audio latent representations can be transformed into waveform signals. Within this framework, a diffusion-based model, denoted as $f_\theta$, is trained to generate audio latent representations conditioned on $\mathbf{Y}$, which typically consists of video frames (and text) in standard video-to-audio generation scenarios.

We employ a flow matching framework to train the latent diffusion model. Let $\mathbf{x}_1$ denote the latent variable. The model is trained to approximate the conditional flow using the following objective:
\begin{equation}
    \mathcal{L}_{\text{CFM}} =
    \mathbb{E}_{t, q(\mathbf{x}_0), q(\mathbf{x}_1, \mathbf{c})}
    \big\| f_\theta(t, \mathbf{Y}, \mathbf{x}_t) - u(\mathbf{x}_t|\mathbf{x}_0, \mathbf{x}_1) \big\|^2,
\label{eq:cfm_objective}
\end{equation}
where $t \in [0,1]$, $\mathbf{x}_t = (1-t)\mathbf{x}_0 + t \mathbf{x}_1$, and $u(\mathbf{x}_t|\mathbf{x}_0,\mathbf{x}_1) = \mathbf{x}_1 - \mathbf{x}_0$ defines the target flow velocity between the Gaussian source and the data manifold.  
Minimizing Eq.~(\ref{eq:cfm_objective}) encourages $f_\theta$ to smoothly transport samples from the prior toward realistic audio latents under the given condition $\mathbf{Y}$. The diffusion model enables the generation of $\mathbf{x}_1 \in \mathbb{R}^d$ from randomly sampled Gaussian noise $\mathbf{x}_0 \in \mathbb{R}^d$, by utilizing the learned time-dependent velocity vector field $f_\theta(t, \mathbf{Y}, \mathbf{x}_t)$.

\vspace{0.5em}
\noindent\textbf{Our approach for incorporating physical cues.}
We define $\mathbf{Y}$ as a set of multiple modalities, encompassing not only video and text but also physical parameters.  
During the latent generation process, the latents evolve from $\mathbf{x}_0$ to $\mathbf{x}_1$ through stacked diffusion transformer (DiT)~\cite{peebles2023scalable} blocks that propagate cross-modal information from $\mathbf{Y}$.
In our approach, all conditional modalities—visual, textual, and physical—are projected into a unified hidden space through lightweight encoders and are synchronized across diffusion timesteps using positional embeddings.  
This backbone provides a stable generative trajectory and serves as the foundation for introducing our physics-aware conditioning. 

\color{black}

\subsection{Overview}
\label{sec:overview}

\begin{figure*}
    \centering
    \includegraphics[width=\linewidth]{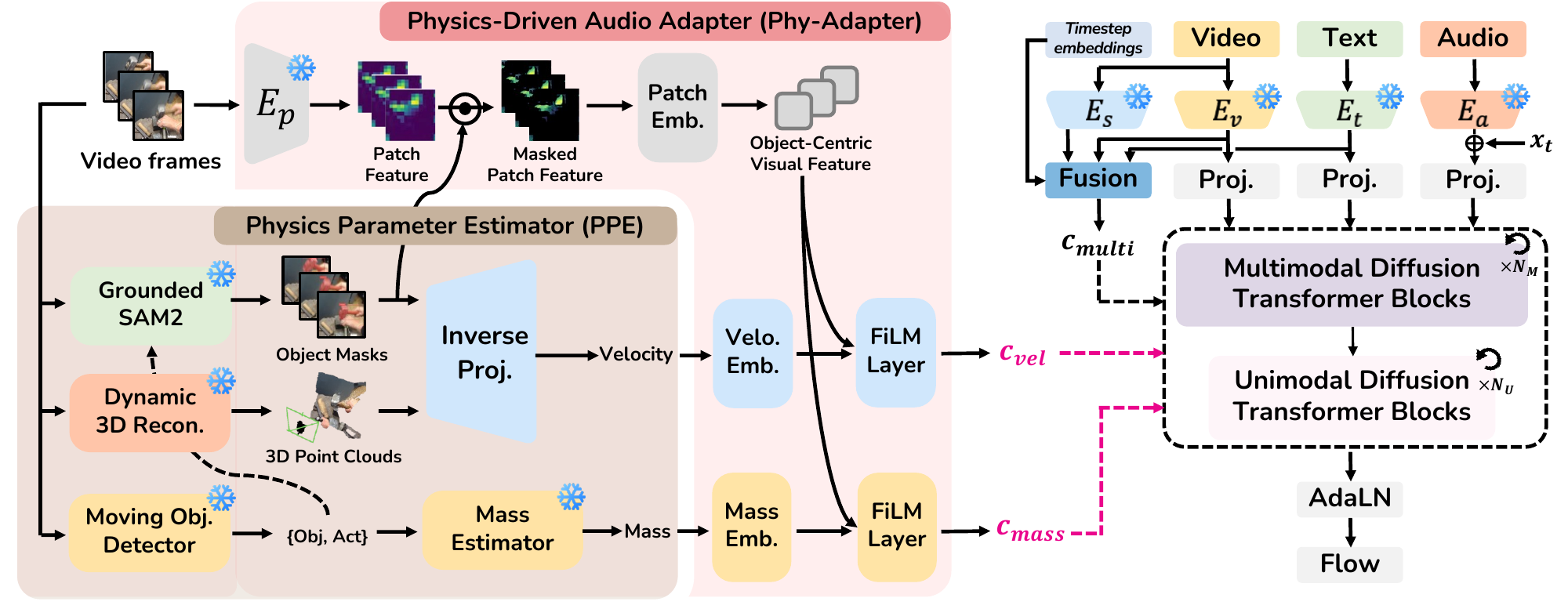}
    \caption{
    \textbf{Overall pipeline of the proposed Physics-Aware Video-to-Audio Synthesis (PAVAS).}
    Given an input video, the Physics Parameter Estimator (PPE) extracts object-level 
    mass and velocity.
    These physics cues are encoded by the Physics-Driven Audio Adapter (Phy-Adapter) and injected into the latent diffusion model alongside multimodal conditions.
    $E_p$ stands for CLIP vision encoder for patch embeddings, $E_s$ for SyncFormer~\cite{iashin2024synchformer} vision encoder, $E_v$ for CLIP vision encoder for flattened visual tokens, $E_t$ for CLIP text encoder, and $E_a$ for VAE/STFT-based audio encoder.
    Dashed lines indicate conditioning pathways, and \color{magenta}{magenta}\color{black}~highlights physics-based conditioning.
    } 
    \label{fig:pipeline}
\end{figure*}

Figure~\ref{fig:pipeline} illustrates the 
overall pipeline of PAVAS.
This leverages a multimodal diffusion transformer architecture~\cite{esser2024scaling,labs2025flux1kontextflowmatching,cheng2025mmaudio}, consisting of transformer blocks shared across visual, textual, and auditory modalities, followed by unimodal blocks specialized for audio decoding.

The \textit{Physics Parameter Estimator} first detects all perceptually moving objects in the input video and estimates object-level, time-invariant mass and per-frame velocities using a combination of vision–language models, segmentation, and dynamic 3D reconstruction. Simultaneously, a vision encoder extracts patch-wise feature embeddings, which are converted into object-centric visual features using masks obtained during velocity estimation. These features provide spatial and semantic context aligned with instance masks, supporting temporal object tracking and facilitating the injection of physical parameters.

Subsequently, the \textit{Physics-Driven Audio Adapter} fuses the physical parameters with object-centric visual features to form temporally aligned, physics-aware representations. To ensure stable and effective fusion, we introduce a $\Delta$\textit{-modulation} mechanism, which enables gradual injection of the fused features into the multimodal transformer blocks of the diffusion backbone. This process guides the model to generate audio that is both perceptually realistic and physically grounded in the underlying object dynamics.

\subsection{Physics Parameter Estimator}
\label{sec:extractor}

The physics parameter estimator quantifies object-wise \textit{mass} and \textit{velocity} from unconstrained videos, providing a quantitative link between visual dynamics and physically grounded sound generation. Given an input video $\{I_\ell\}_{\ell=1}^{L}$, the estimator identifies all moving objects $\mathcal{O}=\{o_i\}_{i=1,2,\ldots}$ and computes, for each object $o_i$, a time-invariant mass $m_i$ (in kilograms) and a time-varying velocity sequence $\{v_i^\ell\}_{\ell=1}^{L-1}$ (in meters per second) in metric 3D space:
\begin{equation}
\mathcal{P} = \big\{ (m_i,\{v_i^\ell\}_{\ell=1}^{L-1}) \,\big|\, o_i \in \mathcal{O} \big\}.
\label{eq:phys_extractor_overview}
\end{equation}
These quantities are estimated through a unified three-stage pipeline:
(i) \textit{Moving-object detection} uses a Vision–Language Model (VLM)~\cite{bai2025qwen2} to localize dynamic entities and generate textual object-level descriptors;
(ii) the same VLM performs \textit{mass estimation} based on visual appearance and textual cues;
(iii) a combination of text-grounded segmentation~\cite{ren2024grounded} and dynamic 3D reconstruction~\cite{wang2025continuous} recovers object-centric geometry and metric-scale velocity.

\vspace{5pt}
\noindent\textbf{Moving-object discovery.}
We first identify entities exhibiting genuine motion, excluding apparent displacement from camera movement.  
A vision-language model~\cite{bai2025qwen2} is prompted with instructions to define “moving objects” (see supplementary material for details). The model outputs a structured set $\mathcal{S} = \{ s_i \}_{i=1}^{N}$, where each entry $s_i$ consists of a localized moving object $o_i$ (\eg, “runner in striped shirt”) and its action $a_i$ (\eg, “sprinting”), \ie, $s_i = (o_i,\, a_i)$. This text-level representation enables open-world generalization and serves as a semantic interface for mass and velocity estimation.

\vspace{5pt}
\noindent\textbf{Mass estimation.}
For each object $o_i$ in $\mathcal{S}$, physical mass is estimated using the Vision-Language Model (VLM)~\cite{bai2025qwen2}.  
Given the object name $o_i$ and action $a_i$, a textual prompt $\mathcal{T}_{\text{mass}}$ is constructed to instruct the model to infer the object’s mass (see supplementary material for details).  
The prompt and video context are provided to the mass estimator $f_{\mathrm{mass}}$, yielding 
$m_i = f_{\mathrm{mass}}\big(I_{1:L},\, \mathcal{T}_{\text{mass}}\big)$.
Unlike geometry-based methods~\cite{standley2017image2mass,nerf2physics} that require multi-view supervision, our approach operates directly on monocular dynamic videos and generalizes across diverse object categories and scenes. The resulting $\{m_i\}_i$ values are time-invariant and later paired with velocity sequences.

\vspace{5pt}
\noindent\textbf{Velocity estimation.}
Reliable velocity estimation for physics-aware conditioning requires metric-scale, temporally coherent per-object trajectories across arbitrary categories.
However, existing monocular depth estimation models~\cite{bhat2023zoedepth,piccinelli2024unidepth,hu2025depthcrafter,yang2024depth,chen2025video} do not meet these requirements: they either lack open-vocabulary, instance-specific object handling or fail to provide temporally consistent metric geometry, and therefore cannot yield physically interpretable object velocities.
To satisfy these requirements, we combine open-vocabulary segmentation with dynamic 3D reconstruction, enabling object-wise centroid trajectories in metric world coordinates for subsequent velocity computation.

Textual object descriptors from $\mathcal{S}$ are used as prompts for Florence-2~\cite{xiao2024florence}, which outputs bounding boxes aligned to target classes. 
SAM-2~\cite{ravi2024sam} refines these boxes into pixel-accurate binary masks and propagates them across time, yielding a sequence of object instance masks $\{\mathbf{M}_i^\ell\}_{\ell=1}^{L}$, where $\mathbf{M}_i^\ell \in \{0,1\}^{H\times W}$.
CUT3R~\cite{wang2025continuous} reconstructs dense 3D geometry for each frame $\ell$, producing a set of 3D points (point cloud) $\mathbf{P}^\ell = \{\mathbf{p}^{\ell,k}\in\mathbb{R}^3\}_{k=1}^{K_\ell}$ and corresponding camera extrinsics $(\mathbf{R}^\ell,\mathbf{T}^\ell)$ under a shared world coordinate system, where $K_\ell$ is the number of point clouds in frame $\ell$. 

Each object’s 3D extent is obtained by inverse-projecting its 2D mask $\mathbf{M}_i^\ell$ onto the reconstructed 3D scene.  
Each pixel $(h,w)$ in $\mathbf{M}_i^\ell$ is mapped to its corresponding 3D point $\mathbf{p}^{\ell,k}$ via the 2D-3D mapping $g_{\mathrm{inv}}$ from CUT3R, aggregating matched points into object-wise 3D points:
$\mathbf{X}_i^\ell 
= g_{\mathrm{inv}}(\mathbf{M}_i^\ell, \mathbf{P}^\ell)
= \{\mathbf{p}^{\ell,k}\in\mathbf{P}^\ell 
\mid \mathbf{M}_i^\ell(h,w)=1\}$, where $k, h,$ and $w$ range over $1 \leq k \leq K_\ell,\, 1 \leq h \leq H,$ and $1 \leq w \leq W$. The centroid $\mathbf{c}_i^\ell$ is then computed to specify the spatial position in metric 3D space for each object: $\mathbf{c}_i^\ell = \frac{1}{|\mathbf{X}_i^\ell|}\sum_{\mathbf{x}\in\mathbf{X}_i^\ell}\mathbf{x}.$

Given the video frame rate $\mathrm{FPS}$ and $\Delta \tau = 1/\mathrm{FPS}$, displacement and instantaneous velocity are computed as
\begin{equation}
d_i^\ell = \|\mathbf{c}_i^{\ell+1}-\mathbf{c}_i^\ell\|_2,
\qquad
v_i^\ell = d_i^\ell / \Delta \tau.
\label{eq:velocity_main}
\end{equation}
The resulting velocity sequence $\mathbf{v}_i$ provides a metric-scale trajectory that captures object-level temporal motion for subsequent physical conditioning.

\subsection{Physics-Driven Audio Adapter}
\label{sec:phyda}
\textit{Physics-Driven Audio Adapter (Phy-Adapter)} bridges the physical values estimated in Sec.~\ref{sec:extractor} with the multimodal latent diffusion backbone, transforming per-object mass and velocity into temporally aligned conditioning signals.
It receives three inputs at each frame $\ell$: 
(i) visual patch embedding $\mathbf{V}^\ell$ from a CLIP-ViT~\cite{ilharco2021openclip} encoder, 
(ii) binary mask $\mathbf{M}_i^\ell$ for each object from Florence-2~\cite{xiao2024florence} and SAM2~\cite{ravi2024sam}, 
and (iii) physical values $\{m_i,v_i\}$ estimated from our Physics Parameter Estimator.
From these, Phy-Adapter constructs physics conditions that are aligned with the visual feature sequence and include all the moving object information.
Then, we inject them into the diffusion transformer’s conditioning stream.
This integration guides the diffusion trajectory toward physically consistent audio generation.

\vspace{4pt}
\noindent\textbf{Object feature extraction.}
The first stage of Phy-Adapter converts raw visual patch embeddings into object-conditioned representations that serve as the base for subsequent physical modulation.
For each frame $\ell$ and detected object $o_i$, the module localizes the object region within the visual feature map $\mathbf{V}^\ell\!\in\!\mathbb{R}^{H\times W\times D_p}$ using its binary segmentation mask $\mathbf{M}_i^\ell\!\in\!\{0,1\}^{H\times W}$ obtained from Florence-2~\cite{xiao2024florence} and SAM2~\cite{ravi2024sam}.
An object-specific feature per frame is computed via masked summation: 
\begin{equation}    
\mathbf{f}_i^\ell = \sum_{h,w} \mathbf{M}_i^\ell[h,w]\cdot\mathbf{V}^\ell[h,w,:],
\end{equation}
where each patch embedding $\mathbf{V}^\ell[h,w,:]\!\in\!\mathbb{R}^{D_p}$ encodes local appearance and motion cues.
The aggregated vector $\mathbf{f}_i^\ell$ is subsequently projected into a hidden space via an affine linear transformation, followed by the application of a lightweight LayerNorm, yielding
$\mathbf{h}_i^\ell \in \mathbb{R}^{D_h}$.
Frames where the object is absent or occluded are replaced by a learnable \emph{object-occlusion token} 
$\mathbf{z}_{\mathrm{obj\text{-}occ}}\!\in\!\mathbb{R}^{D_h}$,
ensuring temporal continuity and stability in the object-level conditioning stream.
The resulting set $\{\mathbf{h}_i^\ell\}_{i,\ell}$ provides object-centric spatio-temporal context upon which mass and velocity modulation are subsequently applied.

\vspace{4pt}
\noindent\textbf{Mass and velocity modulation.}
To condition object-centric visual features on physical properties, both mass and velocity are processed through a similar pipeline. Scalar mass $m_i$ is normalized via $\log(1+m_i)$ and $z$-score normalization with dataset statistics $(\mu_{\text{mass}},\sigma_{\text{mass}})$, while velocity $v_i^\ell$ is $z$-normalized using $(\mu_{\text{vel}}, \sigma_{\text{vel}})$. Each normalized value is expanded using \textit{Fourier feature mapping}~\cite{tancik2020fourier}:
\begin{align}
&\mathbf{f}_{\text{mass},i} = [\,\sin(2\pi\omega_k m_i),\, \cos(2\pi\omega_k m_i)\,]_{k=1}^{K} \in \mathbb{R}^{2K}, \\
&\mathbf{f}_{\text{vel},i}^\ell = [\,\sin(2\pi\omega_k v_i^\ell),\, \cos(2\pi\omega_k v_i^\ell)\,]_{k=1}^{K} \in \mathbb{R}^{2K}.
\end{align}
These features are projected by separate MLPs into embeddings $\mathbf{e}_{\text{mass},i}$ and $\mathbf{e}_{\text{vel},i}^\ell$ in $\mathbb{R}^{D_h}$. Subsequently, we broadcast $\mathbf{e}_{\text{mass},i}$ and flatten $\mathbf{e}_{\text{vel},i}^\ell$ to reshape them into $\tilde{\mathbf{e}}_{\text{mass},i}\in\mathbb{R}^{LD_h}$ and flatten $\tilde{\mathbf{e}}_{\text{vel},i}\in\mathbb{R}^{LD_h}$, respectively. 
FiLM \citep{perez2018film} coefficients are then generated as $(\gamma_{\text{mass},i}, \beta_{\text{mass},i}) = \mathrm{Linear}(\tilde{\mathbf{e}}_{\text{mass},i})$ and $(\gamma_{\text{vel},i}, \beta_{\text{vel},i}) = \mathrm{Linear}(\tilde{\mathbf{e}}_{\text{vel},i})$, and applied to modulate the object-centric visual feature $\mathbf{h}_i$:
\begin{align}
&\mathbf{h}_{\text{mass},i} =\! (1\!+\!\tfrac{1}{2}\!\tanh(\gamma_{\text{mass},i})) \odot \mathbf{h}_i + \tfrac{1}{2}\!\tanh(\beta_{\text{mass},i}), \\
&\mathbf{h}_{\text{vel},i} = (1+\tfrac{1}{2}\tanh(\gamma_{\text{vel},i}))\odot\mathbf{h}_i + \tfrac{1}{2}\tanh(\beta_{\text{vel},i}).
\end{align}
Mass modulation remains constant across time for each object, controlling global loudness and decay, while velocity modulation is frame-dependent, adapting audio features to instantaneous motion. For frames where the object is absent or occluded, a learnable \emph{velocity-occlusion token} $\mathbf{z}_{\mathrm{vel\text{-}occ}}\!\in\!\mathbb{R}^{D_h}$ ensures temporal consistency in velocity conditioning.
\color{black}

\vspace{4pt}
\noindent\textbf{Aggregation and physics-aware conditioning.}
Finally, the modulated mass and velocity features are aggregated via gated pooling:
\begin{equation}
\mathbf{c}_{\text{mass}}=\frac{\sum_i G_{\text{mass},i}\mathbf{h}_{\text{mass},i}}{\sum_i G_{\text{mass},i}},\quad
\mathbf{c}_{\text{vel}}=\frac{\sum_i G_{\text{vel},i}\mathbf{h}_{\text{vel},i}}{\sum_i G_{\text{vel},i}},
\end{equation}
where $G_{\text{mass},i},G_{\text{vel},i}=\sigma(\mathrm{MLP}(\mathbf{h}_{\text{mass},i})),\sigma(\mathrm{MLP}(\mathbf{h}_{\text{vel},i}))$ and $\sigma$ denotes a sigmoid function. 
We omit the frame index $\ell$ from variables here for brevity.

We construct a multimodal condition $\mathbf{c}_{\text{multi}}$ by fusing visual features from the CLIP visual encoder~\cite{ilharco2021openclip}, synchronization features from Synchformer~\cite{iashin2024synchformer}, text embeddings from the CLIP text encoder, and timestep embeddings, following MMAudio~\cite{cheng2025mmaudio}.
This multimodal condition is injected into each transformer block through Adaptive Layer Normalization (AdaLN) layers, providing a unified global and dynamic audiovisual context.

To incorporate physical cues while preserving multimodal stability, we introduce a \textit{$\Delta$-modulation} mechanism that augments AdaLN coefficients with zero-initialized residual mixers driven by physics conditions.
Rather than directly summing physics features with multimodal conditions, each transformer block refines its modulation parameters $\omega$ as
\vspace{-3mm}
\begin{equation}
\tilde{\omega} = \omega(\mathbf{c}_{\text{multi}}) 
\,{\color{magenta}+\,\alpha_m\,g_m(\mathbf{c}_{\text{mass}}) + \alpha_v\,g_v(\mathbf{c}_{\text{vel}})},
\end{equation}
where $g_m,g_v$ are lightweight zero-initialized MLPs and $\alpha_m,\alpha_v$ are learnable gates controlling their magnitude.
$\omega$ denotes the AdaLN modulation parameters (scale and shift) computed from the multimodal condition, and $\tilde{\omega}$ represents their physics-augmented counterpart used within each transformer block.
This residual formulation ensures that physical cues are injected gradually—allowing the model to adapt mass and motion effects without perturbing the multimodal representation—thereby aligning diffusion dynamics with physically consistent audiovisual behavior (see supplementary material for details on the diffusion transformer blocks).

\section{Experiments}
We first outline the evaluation setup, and then present thorough analyses of the experimental results.
Due to the space limitation, we present more implementation details and additional experiments in the supplementary material.

\subsection{Experimental Settings}

\paragraph{Implementation details.}
PAVAS is trained in two stages. In the first stage, we train a multimodal latent diffusion backbone~\cite{esser2024scaling,labs2025flux1kontextflowmatching,cheng2025mmaudio} for general video-to-audio generation.
Then, we integrate the Physics Parameter Estimator (PPE) and the Physics-Driven Audio Adapter (Phy-Adapter) into the backbone to inject explicit mass and velocity cues.
The audio, visual, and text encoders are frozen while the diffusion transformer blocks and conditioning pathways are optimized.
The backbone is trained for 300k iterations using AdamW (learning rate $1{\times}10^{-4}$, weight decay $1{\times}10^{-6}$) with gradient clipping and a batch size of~512. 
The same optimization settings are used in the second stage, except that the number of iterations and the learning rate are reduced to 30K and $1{\times}10^{-5}$, respectively.

During physics-aware fine-tuning, physics tokens are replaced with their corresponding empty tokens with probability 0.1 to handle cases where motion cues are missing or not detected.
\hboh{PAVAS does not strictly assume perfectly trackable visible objects: learnable occlusion tokens handle missing object and velocity cues, while text conditioning still enables the generation of off-screen sound sources.}
We train model variants operating at 16kHz and 44.1kHz, which differ only in backbone capacity.

\paragraph{Dataset.}
We train models using a combination of multimodal and audio–text corpora.
For general video-to-audio learning, we draw video–audio pairs from VGGSound~\cite{chen2020vggsound} and complement them with large-scale audio–text datasets~\cite{kim2019audiocaps,drossos2020clotho,mei2024wavcaps}.
For audio–text datasets, the missing visual tokens are replaced with learnable tokens, allowing the multimodal backbone to remain compatible across data sources.
When training the physics-aware components in the second stage, only the VGGSound dataset is used.
Audio clips from all datasets are clipped to 8-second segments.
Following common practice~\cite{jeong2025read,liu2024tell,cheng2025mmaudio}, we feed the corresponding VGGSound class label as a text input of the model.

\hboh{To assess physical realism, we additionally curate VGG-Impact,
a subset of the VGGSound test split composed of 10 sound classes
and 272 impact moments, where object mass and motion directly
influence the produced sound, such as \textit{hammering nails} and
\textit{basketball bounce}. We first filter clips based on sound
classes. We then manually remove ambiguous scenes with poorly
defined contact dynamics and retain momentary interactions where
impact strength and object inertia are visually identifiable. This
enables targeted evaluation of whether generated audio reflects the
underlying physical parameters inferred from video.}

\renewcommand{\arraystretch}{0.95}
\begin{table*}
\small
    \centering
    \resizebox{1.0\linewidth}{!}{
    \begin{NiceTabular}{l@{\hspace{4pt}}c@{\hspace{8pt}}c@{\hspace{4pt}}c@{\hspace{4pt}}c@{\hspace{4pt}}c@{\hspace{4pt}}c@{\hspace{4pt}}c@{\hspace{1pt}}c@{\hspace{1pt}}c@{\hspace{1pt}}c}
    \toprule
    Method 
    & Params 
    & {\footnotesize Physics corr.}
    & \multicolumn{5}{c}{{\footnotesize Distribution matching}} 
    & {\footnotesize Audio quality} 
    & {\footnotesize Semantic align.} 
    & {\footnotesize Temporal align.} \\
    \cmidrule(lr{\dimexpr 4\tabcolsep-16pt}){3-3}
    \cmidrule(lr{\dimexpr 4\tabcolsep-16pt}){4-8}
    \cmidrule(lr{\dimexpr 4\tabcolsep-16pt}){9-9}
    \cmidrule(lr{\dimexpr 4\tabcolsep-16pt}){10-10}
    \cmidrule(lr{\dimexpr 4\tabcolsep-16pt}){11-11}
    &  
    & APCC-$\Delta$\(\downarrow\)
    & \fdpasst$\downarrow$ 
    & \fdpann$\downarrow$ 
    & \fdvgg$\downarrow$ 
    & \klpann$\downarrow$ 
    & \klpasst$\downarrow$ 
    & \ispann$\uparrow$ 
    & IB-score$\uparrow$ 
    & DeSync$\downarrow$ \\
    \midrule

    See \& Hear~\cite{xing2024seeing}$^\ast$ 
    & 415M & 0.566 
    & 219.0 & 24.58 & 5.40 & 2.26 & 2.30 
    & 8.58 & 33.99 & 1.204 \\

    V-AURA~\cite{viertola2025temporally}$^{\ast\lozenge}$ 
    & 695M  & 0.654
    & 218.5 & 14.80 & 2.88 & 2.42 & 2.07 
    & 10.08 & 27.64 & 0.654 \\

    VATT~\cite{liu2024tell}$^\dagger$ 
    & - & 0.673
    & 131.9 & 10.63 & 2.77 & \textbf{1.48} & 1.41 
    & 11.90 & 25.00 & 1.195 \\

    Frieren~\cite{wang2025frieren}$^{\dagger\lozenge}$ 
    & 159M  & 0.662
    & 106.1 & 11.45 & 1.34 & 2.73 & 2.86 
    & 12.25 & 22.78 & 0.851 \\

    FoleyCrafter~\cite{zhang2024foleycrafter}$^\ast$ 
    & 1.22B & 0.588
    & 140.1 & 16.24 & 2.51 & 2.30 & 2.23 
    & 15.68 & 25.68 & 1.225 \\

    V2A-Mapper~\cite{wang2024v2a}$^{\dagger\lozenge}$ 
    & 229M & 0.671
    & 84.57 & 8.40 & \textbf{0.84} & 2.69 & 2.56 
    & 12.47 & 22.58 & 1.225 \\

    TARO~\cite{ton2025taro}$^{\ast\lozenge}$ 
    & 258M & 0.758
    & 159.1 & 10.49 & 1.57 & 2.92 & 2.67 
    & 9.62 & 22.85 & 1.169 \\



    MMAudio-L~\cite{cheng2025mmaudio}$^\dagger$
    & 1.03B & 0.536
    & 60.60 & 4.72 & 0.97 & 1.65 & 1.40 
    & 17.40 & 33.22 & \textbf{0.442} \\



    \rowcolor{defaultColor}
    PAVAS-L (Ours)
    & 1.04B & \textbf{0.378}
    & \textbf{47.38} & \textbf{3.99} & 1.15 & 1.55 & \textbf{1.35} 
    & \textbf{17.51} & \textbf{35.41} & 0.446 \\

    \bottomrule
    \end{NiceTabular}
    }
    \vspace{0.5mm}
    \caption{\textbf{Quantitative comparison on the VGGSound test set.}
    Following the standard evaluation protocol~\cite{wang2025frieren,cheng2025mmaudio}, parameter counts exclude pretrained encoders (\eg, CLIP), latent audio encoders/decoders, and the modules that are not used in test time (\eg, vocoders).
    We report only the Large variants of MMAudio and PAVAS; both operate at 44.1kHz, while all other models run at 16kHz.
    $\ast$: results reproduced using publicly released code.
    $\dagger$: results evaluated from author-provided samples.
    $\lozenge$: models that do not use text input at test time.}
    \vspace{1mm}
    \label{tab:main_results}
\end{table*}

\paragraph{Metrics.}
We evaluate generation performance across established dimensions—distributional fidelity, perceptual quality, semantic alignment, and temporal synchronization.  
Following prior work~\cite{iashin2021taming,wang2025frieren,cheng2025mmaudio},  
we report Fréchet Distance (FD) and Kullback–Leibler divergence (KL) to measure feature distribution similarity between generated and ground-truth audio,  
using PaSST~\cite{koutini2021efficient}, PANNs~\cite{kong2020panns}, and VGGish~\cite{gemmeke2017audio}.  
Audio quality is measured using the Inception Score (IS).
Semantic alignment is evaluated via cross-modal cosine similarity in the ImageBind space~\cite{girdhar2023imagebind}.
Temporal synchronization is quantified using the DeSync metric from Synchformer~\cite{iashin2024synchformer}, which estimates audio–video misalignment.

To assess the physical plausibility of generated audio, we introduce the \textit{Audio--Physics Correlation Coefficient (APCC)}, which measures how changes in physical magnitude relate to changes in acoustic onset strength.
For each impact event, we first detect the onset in the audio and measure its spectral strength using a standard onset detection function~\cite{Boeck2013}. 
We then estimate the change of kinetic energy at the onset time using the object’s predicted mass and its pre- and post-impact velocities in the corresponding video.
This quantity represents the mechanical energy lost at impact, expected to be radiated as an acoustic impulse~\cite{kinsler2000fundamentals}. 
APCC measures the correlation between these two sequences, kinetic energy changes and acoustic onset strengths, within each sound class in VGG-Impact. We compute this correlation for both ground-truth and generated audio and compare the two. A lower APCC-$\Delta$ indicates that the generated audio more closely matches the real coupling between kinetic energy changes and spectral onset strength, and thus exhibits stronger physics consistency.
\hboh{Detailed definitions and analyses of APCC are provided in the supplementary material.}

\subsection{Experimental Results and Analyses}
\label{sec:results}

We evaluate our framework on both perceptual and physical dimensions, examining (i) how existing video-to-audio generation models behave under physically grounded evaluation, 
(ii) how PAVAS improves perceptual and physical quality, and (iii) which design factors contribute to its effectiveness.

\vspace{4pt}
\noindent\textbf{How well do existing models reflect physics?}
To analyze whether current Video-to-Audio (V2A) models generate acoustics that follow real physical behavior, we evaluate nine state-of-the-art Video-to-Audio (V2A) models~\cite{xing2024seeing,viertola2025temporally,liu2024tell,wang2025frieren,zhang2024foleycrafter,wang2024v2a,ton2025taro,cheng2025mmaudio}, including ours, on the VGG-Impact benchmark using the proposed Audio–Physics Correlation Coefficient (APCC) (See Table~\ref{tab:main_results}).
Across models, we observe APCC-$\Delta$ values frequently exceeding 0.5, indicating noticeable gaps with respect to the ground-truth correlation between physical quantities and acoustic responses.
PAVAS obtains the lowest APCC-$\Delta$ among evaluated methods, suggesting that it more closely matches the physics–audio relationship present in the dataset.
Overall, the results indicate that while existing V2A models may capture semantic alignment, their generated audio only partially captures variations in underlying physical magnitudes.

\begin{figure*}
    \centering
    \includegraphics[width=0.9\linewidth]{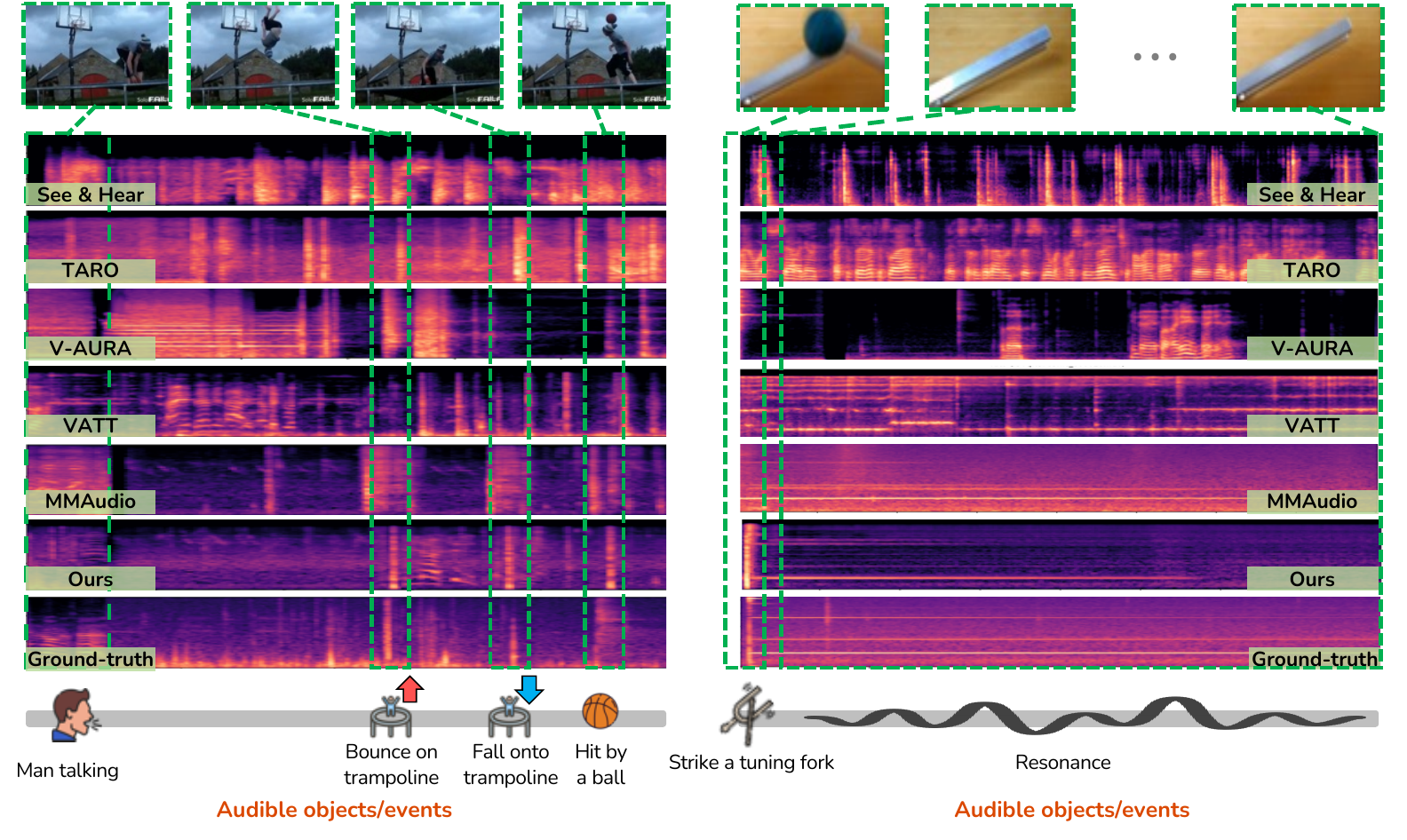}
    \vspace{1mm}
    \caption{
    \textbf{Qualitative comparison of generated spectrograms.}
    We visualize spectrograms from existing V2A models~\cite{xing2024seeing,ton2025taro,viertola2025temporally,liu2024tell,cheng2025mmaudio}, our method, and the ground truth.
    Green dashed lines indicate spectral patterns temporally aligned with visual events in the video, and graphic icons denote audible objects or interactions present in the audio track.
    PAVAS produces spectral patterns that more closely align with these events, whereas other methods often generate components that are not well aligned with the visual dynamics.
    }
    \label{fig:qual}
\end{figure*}

\hboh{This motivates incorporating explicit object-level mass and velocity as conditioning signals to promote more physically consistent V2A generation.
Since such conditioning is only effective when the underlying physical estimates are sufficiently plausible, we assess both components of the Physics Parameter Estimator (PPE).
Across these evaluations, both estimators demonstrate favorable performance, supporting their suitability as conditioning inputs; details and results are provided in the supplementary material.
}

\noindent\textbf{Does PAVAS improve audio generation quality?}
Table~\ref{tab:main_results} presents quantitative comparisons across distributional, semantic, and synchronization metrics against state-of-the-art Video-to-Audio (V2A) models.
PAVAS achieves consistently favorable performance across all measures, suggesting that incorporating explicit physics cues can enhance not only physical plausibility but also audio quality on in-the-wild video data~\cite{chen2020vggsound}.
Spectrogram analysis in Fig.~\ref{fig:qual} reveals that existing V2A models often fail to reflect the impact dynamics present in the video (\eg, a trampoline bounce or a tuning-fork strike).
In contrast, our method generates physically plausible audio whose spectral patterns better reflect the strength and dynamics of the impact, showing a closer match to the ground-truth spectrogram at the moment of interaction.
We further conduct a user study on the VGGSound~\cite{chen2020vggsound} test set, where participants rate four aspects of the generated audio: audio quality, semantic alignment, temporal alignment, and \textit{physical plausibility}. Table~\ref{tab:user_study} shows that PAVAS achieves favorable subjective scores across all criteria.

\renewcommand{\arraystretch}{1.0}
\begin{table}[t]
\centering
\resizebox{1.0\linewidth}{!}{
\begin{NiceTabular}{lcccc}
\toprule
Method 
& Audio qual.\(\uparrow\) 
& Semantic align.\(\uparrow\) 
& Temporal align.\(\uparrow\) 
& Physical plau.\(\uparrow\) \\
\midrule
See \& Hear~\cite{xing2024seeing} 
    & 1.77\pmnum{0.84} 
    & 1.71\pmnum{1.02} 
    & 1.49\pmnum{0.79}
    & 1.63\pmnum{0.92} \\
V-AURA~\cite{viertola2025temporally} 
    & 2.68\pmnum{1.14} 
    & 2.90\pmnum{1.30} 
    & 2.86\pmnum{1.28}
    & 2.58\pmnum{1.24} \\
VATT~\cite{liu2024tell} 
    & 2.27\pmnum{0.93} 
    & 2.51\pmnum{1.20} 
    & 2.00\pmnum{0.98}
    & 2.22\pmnum{1.09} \\
V2A-Mapper~\cite{wang2024v2a} 
    & 2.75\pmnum{1.09} 
    & 2.55\pmnum{1.35} 
    & 1.92\pmnum{0.99}
    & 2.12\pmnum{1.17} \\
TARO~\cite{ton2025taro} 
    & 2.44\pmnum{1.07} 
    & 2.42\pmnum{1.25} 
    & 1.93\pmnum{0.98}
    & 2.08\pmnum{1.09} \\
MMAudio-L~\cite{cheng2025mmaudio} 
    & 3.98\pmnum{0.98} 
    & 4.14\pmnum{1.01} 
    & 4.06\pmnum{1.02}
    & 3.90\pmnum{1.11} \\
\rowcolor{defaultColor}
PAVAS-L (Ours) 
    & \textbf{4.23\pmnum{0.77}} 
    & \textbf{4.47\pmnum{0.71}} 
    & \textbf{4.45\pmnum{0.80}}
    & \textbf{4.37\pmnum{0.84}} \\
\bottomrule
\end{NiceTabular}
} 
\caption{\textbf{User study on the VGGSound test set.}
27 participants rate eight generated audios on four aspects: audio quality, semantic alignment, temporal alignment, and physical plausibility. We report the mean and standard deviation of the Likert~\cite{likert1932technique} scale scores (1--5; strongly disagree, disagree, neutral, agree, strongly agree).
}
\label{tab:user_study}
\vspace{1ex}
\end{table}
\renewcommand{\arraystretch}{0.8}
\begin{table}[t]
\small
\centering
\resizebox{1.0\linewidth}{!}{
\begin{NiceTabular}{lcccc}
\toprule
Setting & \fdpasst\(\downarrow\) & IS\(\uparrow\) & IB-score\(\uparrow\) & DeSync\(\downarrow\) \\
\midrule
\rowcolor{gray!10}\Block[l]{1-5}{\textbf{(A) Additional Adaptation}}\\
Backbone & \textbf{70.19} & \textbf{14.44} & 29.13 & \textbf{0.483} \\
+ Trained longer & 71.99 & 14.34 & \textbf{29.46} & 0.486 \\
\midrule
\rowcolor{gray!10}\Block[l]{1-5}{\textbf{(B) Physics Features (w/ $\Delta$-Modulation)}}\\
+ $\mathbf{c}_\text{mass}$ only & 66.89 & 15.94 & 29.40 & 0.480 \\
+ $\mathbf{c}_\text{vel}$ only & 67.22 & 15.07 & 29.33 & \textbf{0.446} \\
\rowcolor{defaultColor}
+ $\mathbf{c}_\text{mass}$ and $\mathbf{c}_\text{vel}$ (ours) & \textbf{65.67} & \textbf{16.50} & \textbf{29.41} & 0.448 \\
\midrule
\rowcolor{gray!10}\Block[l]{1-5}{\textbf{(C) Injection Strategy (w/ both $\mathbf{c}_\text{mass}$ and $\mathbf{c}_\text{vel}$)}}\\
Direct Summation & 67.31 & 16.30 & 29.40 & 0.455 \\
\rowcolor{defaultColor}
$\Delta$-Modulation (ours)   & \textbf{65.67} & \textbf{16.50} & \textbf{29.41} & \textbf{0.448} \\
\bottomrule
\end{NiceTabular}
}
\vspace{1mm}
\caption{\textbf{Ablation study on additional adaptation, physics features, and injection strategies.}
(A) examines the effect of training the pretrained backbone longer without physics conditions.
(B) investigates the contribution of each physics feature under $\Delta$-modulation.
(C) compares summation vs.\ residual $\Delta$-modulation when both features are used. 
All models use the S-16kHz backbone and share the same training configuration.}
\label{tab:ablation}
\end{table}

\vspace{4pt}
\noindent\textbf{What makes improvements on audio quality?} 
To analyze which components of PAVAS are responsible for the observed improvements in audio generation, we conduct ablation studies summarized in Table~\ref{tab:ablation}.
Simply 
training
the backbone~\cite{cheng2025mmaudio}
longer
on VGGSound~\cite{chen2020vggsound} results in only marginal changes across evaluation metrics, suggesting that improvements in our full model arise from the physics-aware components rather than additional adaptation to the dataset.
In contrast, introducing physics features leads to consistent improvements: conditioning on mass or velocity individually enhances distributional and perceptual metrics, and combining both yields the best results.
We further compare conditioning strategies and find that residual $\Delta$-modulation surpasses direct summation, suggesting that injecting physics cues as residual adaptive modulations within the diffusion transformer allows the model to incorporate physical information more effectively.

\section{Conclusion}
\label{sec:conclusion}
We presented \textbf{PAVAS}, a physics-aware video-to-audio generation method that conditions a latent diffusion model on object-level mass and velocity.
Both quantities are estimated from Physics Parameter Estimator (PPE) and injected via Physics-Driven Audio Adapter (Phy-Adapter), enabling the model to produce sounds whose intensity and temporal structure better reflect underlying visual dynamics.
To evaluate physical plausibility, an aspect overlooked in prior work, we introduce VGG-Impact and Audio–Physics Correlation Coefficient (APCC), which measure how generated audio follows changes in kinetic energy.
Experiments show that mass and velocity conditioning each contribute notable improvements, and their combination yields stronger perceptual quality and better physics consistency than existing models.

\newcommand{\newparagraph}[1]{\noindent \textbf{#1}}
\newparagraph{Discussion \& future direction.}
PAVAS involves several pretrained vision modules.
However, it is not a simple combination of existing components: object-level mass and velocity are inferred without supervised audio–physics labels, and the lightweight Phy-Adapter provides an effective pathway to inject these cues into the diffusion model.
Ablations show that this module contributes gains beyond visual features alone, indicating that physical conditioning is meaningful and complementary. 
Future work may explore more compact adapters, jointly optimized physics estimators, and richer physical factors such as explicit material modeling.
\section{Acknowledgments}
We sincerely thank Dongseok Shim and Koichi Saito for their insightful internal reviews, and Masato Ishii and Takashi Shibuya for their support in reproducing MMAudio.

O. Hyun-Bin and T.-H. Oh was partially supported by the InnoCORE program of the Ministry of Science and ICT(25-InnoCORE-01, Trust-Enhanced Mutualistic Bio-Embedded AI (30\%)),
Institute of Information \& communications Technology Planning \& Evaluation (IITP) grant funded by the Korea government(MSIT) (No.RS-2025-25443318, Physically-grounded Intelligence: A Dual Competency Approach to Embodied AGI through Constructing and Reasoning in the Real World (23.3\%)), 
the National Research Foundation of Korea(NRF) funded by the Korea government(MSIT) (No. RS-2024-00451947 (23.3\%);
No. RS-2024-00358135, Corner Vision: Learning to Look Around the Corner through Multi-modal Signals (23.3\%)).

{
    \small
    \bibliographystyle{ieeenat_fullname}
    \bibliography{main}
}

\clearpage
\maketitlesupplementary

\newtheorem{assume}{Assumption}

\setcounter{section}{0}
\setcounter{figure}{0}
\setcounter{table}{0}
\setcounter{equation}{0}

\renewcommand{\thesection}{\Alph{section}}
\renewcommand{\thefigure}{S\arabic{figure}}
\renewcommand{\thetable}{S\arabic{table}}
\renewcommand{\theequation}{\alph{equation}}

\hypersetup{linkcolor=black}

\section*{Contents}
\noindent\hyperref[sec:A]{\textbf{A \ \ \ Supplementary Video}}\\
\hyperref[sec:B]{\textbf{B \ \ \ Can We Estimate Plausible Physics Values?}} \\
\hyperref[sec:supp_mass]{\text{\quad B.1 \ \ \ Mass Estimation Ability}} \\ 
\hyperref[sec:supp_velocity]{\text{\quad B.2 \ \ \ Velocity Estimation Ability}} \\ 
\hyperref[sec:C]{\textbf{C \ \ \ Robustness to Imperfect Physics Estimates}}\\
\hyperref[sec:supp_sensitivity]{\text{\quad C.1 \ \ \ Sensitivity to Mass Perturbation and Velocity Noise}} \\ 
\hyperref[sec:supp_fallback]{\text{\quad C.2 \ \ \ Graceful Fallback under Missing Physics Cues}} \\ 
\hyperref[sec:D]{\textbf{D \ \ \ Audio-Physics Correlation Benchmark}}\\
\hyperref[sec:supp_vggimpact]{\text{\quad D.1 \ \ \ VGG-Impact}} \\ 
\hyperref[sec:supp_apcc]{\text{\quad D.2 \ \ \ Audio-Physics Correlation Coefficient}} \\ \hyperref[sec:supp_apcc_robustness]{\text{\quad D.3 \ \ \ Robustness of APCC}} \\ 
\hyperref[sec:E]{\textbf{E \ \ \ Module Details}}\\
\hyperref[supp:vlm_prompts]{\text{\quad E.1 \ \ \ Instruction Prompts for VLMs}} \\ 
\hyperref[supp:ppe_config]{\text{\quad E.2 \ \ \ Details of Physics Parameter Estimator}} \\ 
\hyperref[sec:supp_occlusion_token]{\text{\quad E.3 \ \ \ Handling Missing Temporal Observations}} \\ 
\hyperref[app:sec:multimodal_blocks]{\text{\quad E.4 \ \ \ Diffusion Transformer Backbone}} \\ 
\hyperref[app:training]{\text{\quad E.5 \ \ \ Training Details}} \\ 
\hyperref[sec:F]{\textbf{F \ \ \ Runtime and Invalid/Missing-Observation Statistics of PPE}}\\
\hyperref[sec:supp_runtime]{\text{\quad F.1 \ \ \ Module-wise Runtime Profiling}} \\ 
\hyperref[sec:supp_failure]{\text{\quad F.2 \ \ \ Invalid and Missing-Observation Statistics}} \\ 
\hyperref[sec:G]{\textbf{G \ \ \ Generalization Beyond Impact Scenes}} \\
\hyperref[sec:H]{\textbf{H \ \ \ Setup of User Study}} \\
\hyperref[sec:I]{\textbf{I \ \ \ Additional Discussion on Occlusion and Off-Screen Audio}} \\
\hyperref[sec:J]{\textbf{J \ \ \ Additional Qualitative Samples}} \\
\noindent\rule{\linewidth}{0.2pt}

\appendix

\hypersetup{linkcolor=cvprblue}

\section{Supplementary Video}\label{sec:A}
This work focuses on Video-to-Audio (V2A) generation, which is best viewed in video format. 
Please refer to the attached \textbf{supplementary video}.
The video contains qualitative comparisons between two of the latest state-of-the-art V2A models~\cite{cheng2025mmaudio,ton2025taro} and our approach on the VGGSound~\cite{chen2020vggsound} test set, highlighting the perceptual audio quality and physics consistency achieved by PAVAS.
Beyond the generation results on VGGSound test set, the video also includes several controlled generation scenarios using in-the-wild video clips:
\begin{itemize}
\item \emph{Controlled semantics}: a hammer-crashing scene where only the material appearance is changed, demonstrating that PAVAS adjusts the generated impact sound according to the object’s material cues.
\item \emph{Controlled velocity}: manipulating the object’s velocity while keeping mass and material information constant, showing that faster motion produces sharper and higher-energy impact transients.
\item \emph{Controlled mass}: varying the object’s mass while holding its material and motion constant, demonstrating that heavier objects yield proportionally stronger impact sounds.
\end{itemize}
These examples further validate that PAVAS responds coherently to visual and physical cues, producing both perceptually plausible and physically consistent audio.

\section{Can We Estimate Plausible Physics Values?}\label{sec:B}
Reliable physical conditioning requires that the estimated object mass and velocity correspond to plausible real-world magnitudes.
In this section, therefore, we evaluate both components of the Physics Parameter Estimator (PPE) to ensure that the physical values used for conditioning are sufficiently reliable for physics-aware audio synthesis.

\subsection{Mass Estimation Ability}\label{sec:supp_mass}
For mass estimation, we follow the evaluation protocol of NeRF2Physics~\cite{nerf2physics} and test our estimator on 500 objects from the Amazon Berkeley Objects (ABO) dataset~\cite{collins2022abo}.
We report four standard metrics widely used in mass estimation work~\cite{standley2017image2mass,nerf2physics}---Absolute Difference Error (ADE), Absolute Log-Difference Error (ALDE), Absolute Percentage Error (APE), and Minimum Ratio Error (MnRE).
Given ground-truth mass $m$ and predicted mass $\hat{m}$, they are defined as:
\makeatletter
\renewcommand{\theequation}{\arabic{equation}}
\makeatother
\begin{align}
\text{ADE}  &= |m - \hat{m}|, \\
\text{ALDE} &= |\ln m - \ln \hat{m}|, \\
\text{APE}  &= \left|\frac{m - \hat{m}}{m}\right|, \\
\text{MnRE} &= \min\!\left(\frac{m}{\hat{m}},\, \frac{\hat{m}}{m}\right).
\end{align}

Table~\ref{tab:mass} shows that our method achieves favorable performance across four metrics while using only a single-view input, unlike NeRF2Physics which requires multi-view observations.
Moreover, although prior work~\cite{nerf2physics} reported limited generalization when applying earlier Vision–Language Models (VLMs)~\cite{liu2023visual} to physical reasoning, our results show that a modern VLM~\cite{bai2025qwen2}, when guided with carefully designed prompts, can serve as an effective mass estimator.
See Sec.~\ref{supp:vlm_prompts} for an example of a detailed prompt for the mass estimation.
\begin{table}
\small
    \centering
    \resizebox{1.0\linewidth}{!}{
    \begin{NiceTabular}{l@{\hspace{8pt}}c@{\hspace{8pt}}c@{\hspace{8pt}}c@{\hspace{8pt}}c}
    \toprule
    Method & ADE $\downarrow$ & ALDE $\downarrow$ & APE $\downarrow$  & MnRE $\uparrow$ \\
    \midrule
    Image2mass~\cite{standley2017image2mass} & 12.496 & 1.792 & 0.976 & 0.341 \\
    2D CNN & 15.431  & 1.609 & 14.459  & 0.362 \\
    LLaVA~\cite{liu2023visual} & 17.328 & 1.893 & 1.837  & 0.306 \\
    NeRF2Physics~\cite{nerf2physics} & 8.730 & \textbf{0.771} & 1.061 & \textbf{0.552} \\
    \midrule
    \rowcolor{defaultColor}
    Ours & \textbf{6.954} & 0.809  & \textbf{0.823} & 0.529 \\
    \bottomrule
    \end{NiceTabular}
    }
    \caption{\textbf{Mass estimation on ABO-500 test set.} We follow the same evaluation protocol of prior work~\cite{standley2017image2mass,nerf2physics}. Note that NeRF2Physics~\cite{nerf2physics} requires multi-view images to estimate mass.}
    \label{tab:mass}
\end{table}

\begin{figure}
    \centering
    \includegraphics[width=1.0\linewidth]{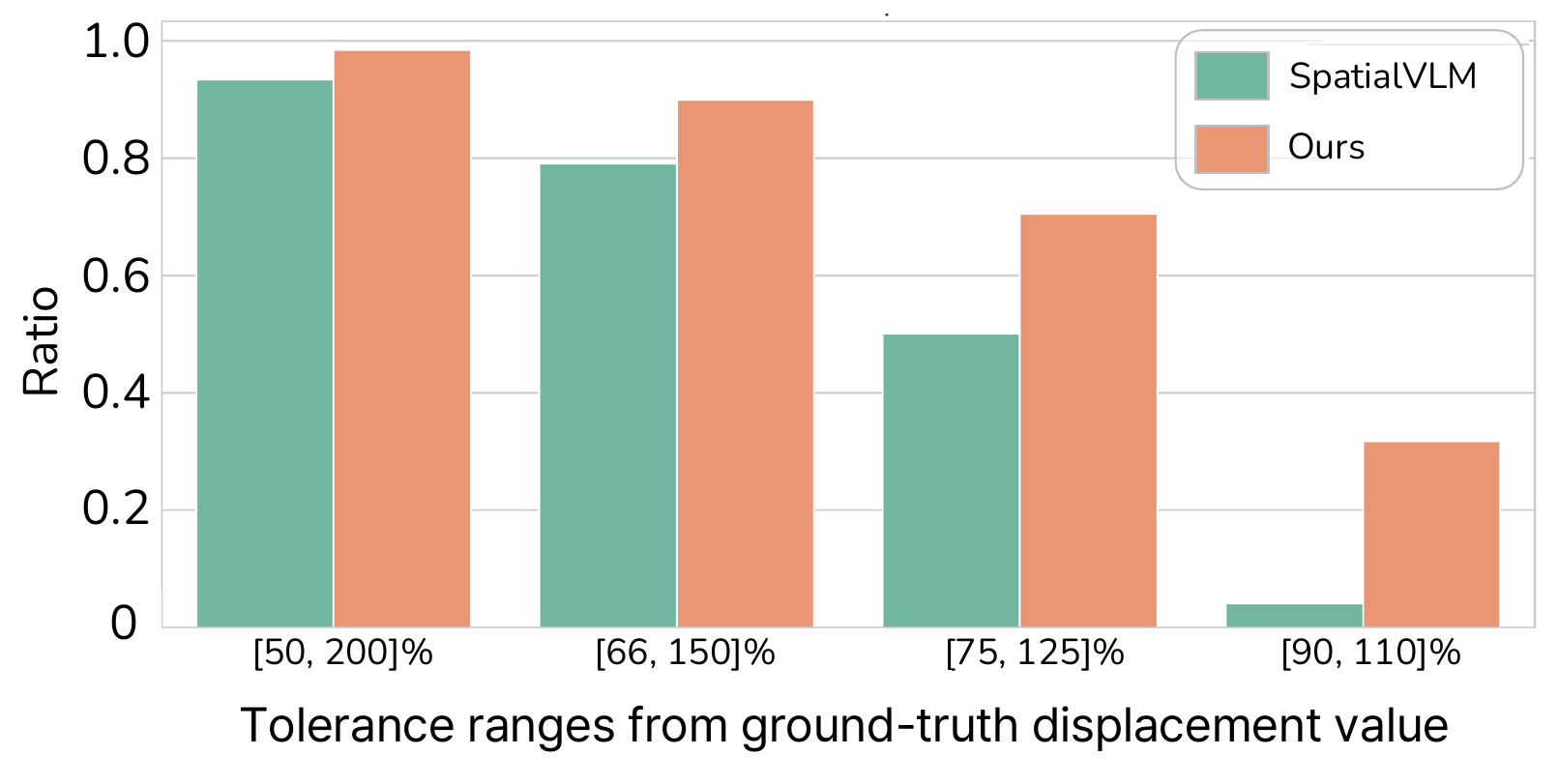}
    \caption{\textbf{Velocity estimation on STARSS23 test set.}     
    We compare our metric-scale displacement estimation with the pseudo ground-truth dataset construction pipeline of SpatialVLM~\cite{chen2024spatialvlm}. We measure the ratio of predictions falling within various tolerance ranges of the ground-truth displacement, providing a proxy evaluation of object-level velocity estimation accuracy.
    }
    \label{fig:velocity}
\end{figure}

\subsection{Velocity Estimation Ability}\label{sec:supp_velocity}
To evaluate velocity estimation performance, we utilize STARSS23~\cite{shimada2023starss23}, which provides metric-depth (i.e., displacement) annotations for sounding and moving objects. We adopt this dataset as a proxy benchmark since no established dataset exists for open-vocabulary object velocity estimation.
We first convert each equirectangular video in STARSS23 into a normal field-of-view format using gnomonic projection~\cite{weisstein2025gnomonic}. Grounded-SAM2~\cite{ravi2024sam} is then applied with the dataset’s sound-class labels to obtain text-conditioned object masks. Low-quality segmentations are manually filtered out, resulting in 256 ten-second video clips. Using these filtered videos and masks, our dynamic 3D reconstruction model combined with inverse projection and centroid aggregation recovers metric-scale 3D trajectories for each object.

To assess the reliability of our metric-scale velocity values, we compare our pipeline not with the reasoning capability of SpatialVLM~\cite{chen2024spatialvlm} but with its pseudo ground-truth dataset construction pipeline, which produces more accurate metric-scale displacement values using pretrained expert models. This setup evaluates the quality of metric-scale 3D lifting itself rather than the reasoning or VQA components of SpatialVLM, and it constitutes a more challenging evaluation than comparing against SpatialVLM’s predicted outputs.
We reproduce this pipeline using the open-source implementation officially acknowledged by the authors\footnote{https://github.com/remyxai/VQASynth} and use the same filtered segmentation masks to ensure a fair comparison. We report the proportion of predictions that fall within several tolerance ranges of the ground-truth displacement (e.g., within $[50,200]\%$).
Figure~\ref{fig:velocity} shows that our method achieves higher accuracy across all tolerance ranges, with particularly large improvements under stricter thresholds. These results indicate that our method provides more reliable metric-scale displacement (\ie, velocity) estimation.

\section{Robustness to Imperfect Physics Estimates}
\label{sec:C}
In this section, we analyze how PAVAS behaves when the estimated physical cues are imperfect or partially missing. Since the mass and velocity estimates used in our pipeline are generally reliable under the adopted proxy evaluations, we interpret the following perturbation experiments as stress tests rather than common operating conditions. 
We evaluate both inference-time sensitivity to perturbed physical values and fallback behavior when all physics tokens are replaced with learnable occlusion tokens.

\begin{figure}[t]
    \centering
    \includegraphics[width=\linewidth]{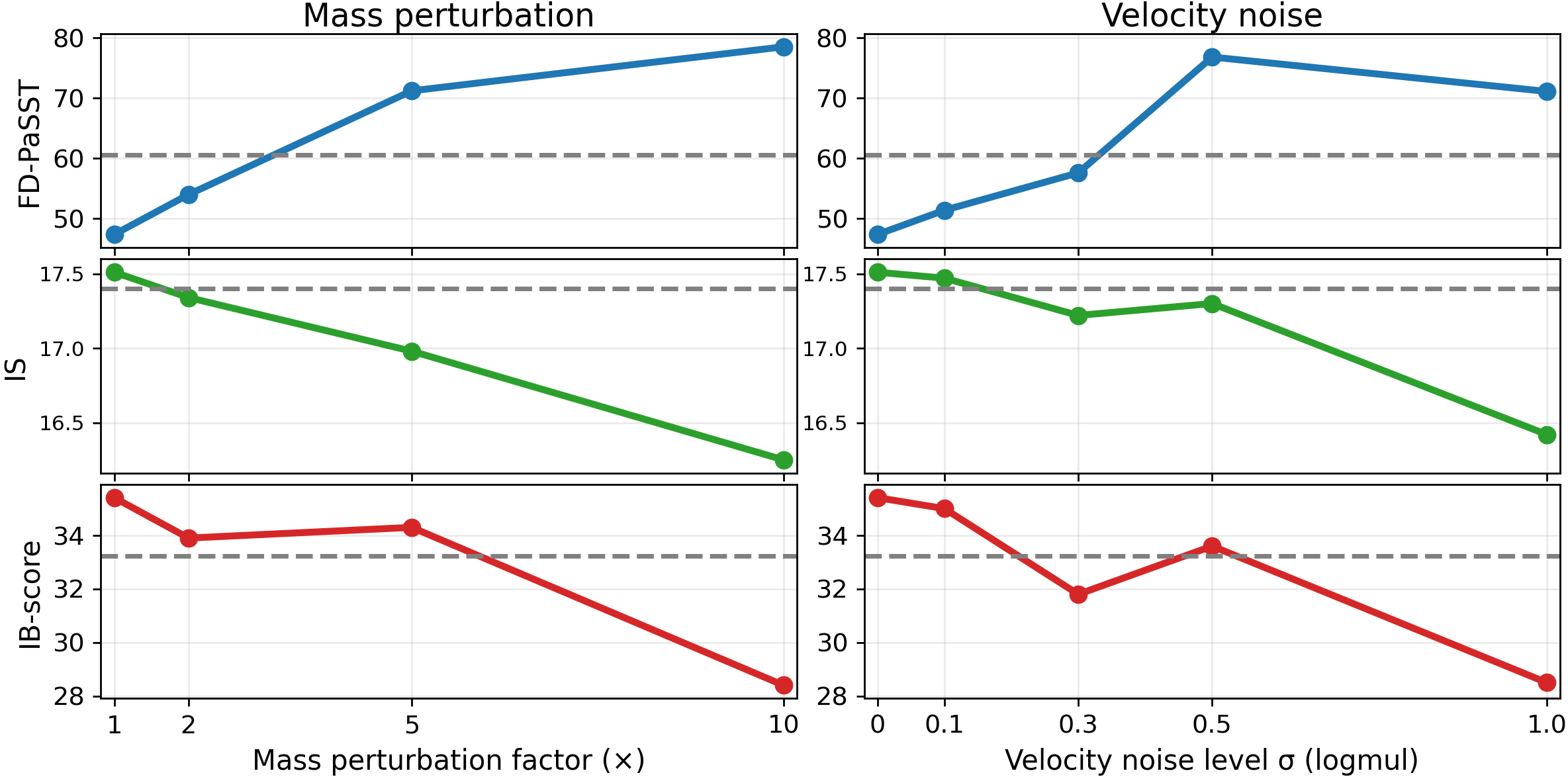}
    \caption{\textbf{Sensitivity of PAVAS-L to perturbed physics estimates at inference time.} \textbf{[Left]} mass perturbation by multiplicative factors $\{2,5,10\}$. \textbf{[Right]} velocity perturbation with log-multiplicative noise $\sigma \in \{0.1,0.3,0.5,1.0\}$. The dashed line denotes MMAudio-L~\cite{cheng2025mmaudio}. PAVAS degrades gradually as the perturbation magnitude increases, showing robustness to moderate estimation errors.}
    \label{fig:sensitivity_noise}
\end{figure}

\subsection{Sensitivity to Mass Perturbation and Velocity Noise}
\label{sec:supp_sensitivity}
To test sensitivity to estimation errors, we perturb the inferred mass and velocity values at inference time. For mass, we multiply the estimated value by factors of $\{2, 5, 10\}$. For velocity, we inject log-multiplicative noise with $\sigma \in \{0.1, 0.3, 0.5, 1.0\}$. Figure~\ref{fig:sensitivity_noise} summarizes the resulting trends, where the dashed line indicates the MMAudio-L~\cite{cheng2025mmaudio}. 
As the perturbation magnitude increases, the performance of PAVAS degrades gradually rather than abruptly, indicating that the proposed conditioning mechanism is robust to moderate estimation errors. 
Noticeable degradation appears only under severe perturbations, such as mass $\times 10$ or velocity noise with $\sigma = 1.0$, which are substantially more extreme than the typical estimation errors observed in practice.


\begin{table}[t]
\centering
\small
\begin{tabular}{lccc}
\toprule
Method & FD$_{\text{PaSST}}\downarrow$ & IS$\uparrow$ & IB$\uparrow$ \\
\midrule
Backbone & 60.6 & 17.4 & 33.2 \\
PAVAS-L & 47.4 & 17.5 & 35.4 \\
\midrule
Fallback & 64.5 & 17.2 & 32.8 \\
\bottomrule
\end{tabular}
\vspace{1mm}
\caption{\textbf{Fallback via occlusion-token replacement.} When all physics tokens are replaced with occlusion tokens at inference time, PAVAS-L degrades gracefully toward the appearance-driven backbone~\cite{cheng2025mmaudio} rather than failing abruptly.}
\label{tab:fallback}
\end{table}

\subsection{Graceful Fallback under Missing Physics Cues}
\label{sec:supp_fallback}
We further analyze the extreme case where valid physical cues are entirely unavailable. In this setting, all mass and velocity tokens are replaced with their corresponding occlusion tokens at inference time, yielding the fallback variant in Table~\ref{tab:fallback}. 
The fallback model performs close to the appearance-driven backbone~\cite{cheng2025mmaudio}, indicating that the proposed design does not fail catastrophically when physics cues are missing. 
Instead, the model gracefully falls back to the underlying multimodal generation pathway. This behavior is consistent with the occlusion-token training strategy used during physics-aware fine-tuning, which encourages robustness to missing or ambiguous motion cues.


\section{Audio-Physics Correlation Benchmark}\label{sec:D}
\subsection{VGG-Impact}\label{sec:supp_vggimpact}
To assess physical realism, we curate VGG-Impact, a subset of the VGGSound test split consisting of 10 impact-related sound classes and 272 impact moments, where object mass and motion directly influence the resulting audio.
The selected sound classes are:

\begin{itemize}
\item \textit{basketball bounce}
\item \textit{bowling impact}
\item \textit{hammering nails}
\item \textit{bouncing on trampoline}
\item \textit{opening or closing car doors}
\item \textit{striking pool}
\item \textit{striking bowling}
\item \textit{forging swords}
\item \textit{chopping wood}
\item \textit{door slamming}
\end{itemize}

We then apply our Physics Parameter Estimator (PPE) to obtain frame-wise object masks and their corresponding physical quantities (mass and velocity). Videos are manually filtered out under two conditions:
(1) when they do not contain clear object–object interaction sounds (we exclude ambiguous scenes with unclear contact dynamics), and
(2) when object masks fail to provide reliable tracks due to occlusion or segmentation errors.

After filtering, we retain video–audio pairs containing identifiable impact events, along with per-object mass and velocity sequences. These serve as the basis for evaluating whether generated audio reflects the underlying physical parameters via the proposed Audio–Physics Correlation Coefficient (APCC).

\subsection{Audio-Physics Correlation Coefficient}\label{sec:supp_apcc}
To assess the physical plausibility of generated audio, we introduce the
\textit{Audio--Physics Correlation Coefficient (APCC)}, which measures how
consistently variations in physical magnitude are reflected in the acoustic
response. We extract impact onsets $\{\tau_j\}$ from the generated waveform
using the SuperFlux onset detector~\cite{Boeck2013}. The onset detection
function (ODF) is computed on the mel-spectrogram $S_{\mathrm{mel}}(\tau,f)$
as a local spectral energy difference:
\begin{equation}
\mathrm{ODF}(\tau)
= \sum_{f}\big[\,S_{\mathrm{mel}}(\tau,f)-S_{\mathrm{mel}}(\tau-\delta,f)\,\big]_+ ,
\end{equation}
where $[\cdot]_+$ denotes half-wave rectification. The spectral onset
energy used for correlation is therefore $y_j=\mathrm{ODF}(\tau_j)$.

To quantify the physical magnitude associated with each event, we measure the
kinetic energy drop for the corresponding object $o_i$ around each onset
$\tau_j$. Given its estimated mass $m_i$ and velocity sequence
$\{v_i^{\ell}\}_{\ell=1}^{L}$, we define the incoming and outgoing velocities
as
\begin{equation}
v_{i,\mathrm{in},j}
=\max_{\ell\in\mathcal{N}^{-}(\tau_j)} v_i^{\ell},
\qquad
v_{i,\mathrm{out},j}
=v_i^{\,\ell_j},
\end{equation}
where $\ell_j$ is the frame index closest to onset time $\tau_j$, and
$\mathcal{N}^{-}(\tau_j)$ denotes a short pre-onset neighborhood. The
corresponding kinetic energy change is
\begin{equation}
\Delta E_{i,j}
=\tfrac{1}{2}\,m_i\!\left((v_{i,\mathrm{in},j})^2-(v_{i,\mathrm{out},j})^2\right),
\end{equation}
representing the mechanical energy lost at impact, which is expected to be
radiated as an acoustic impulse~\cite{kinsler2000fundamentals}.

For each video, we z-normalize $\{\Delta E_{i,j}\}$ and $\{y_j\}$ and compute a
Pearson correlation per interaction class $c$, yielding
$\mathrm{APCC}_{c}$. We evaluate this correlation for both a ground-truth
audio and a model-generated audio, obtaining
$\mathrm{APCC}_{\mathrm{GT},c}$ and $\mathrm{APCC}_{\mathrm{model},c}$,
respectively. The final score is the class-averaged deviation
\begin{equation}
\mathrm{APCC}\text{-}\Delta
=\frac{1}{|\mathcal{C}|}\sum_{c\in\mathcal{C}}
\bigl|\mathrm{APCC}_{\mathrm{GT},c}-\mathrm{APCC}_{\mathrm{model},c}\bigr|.
\end{equation}
A lower APCC-$\Delta$ indicates that the generated audio more closely matches
the real coupling between kinetic energy changes and spectral onset strength.

\subsection{Robustness of APCC}
\label{sec:supp_apcc_robustness}
\begin{figure}[t]
    \centering
    \includegraphics[width=0.8\linewidth]{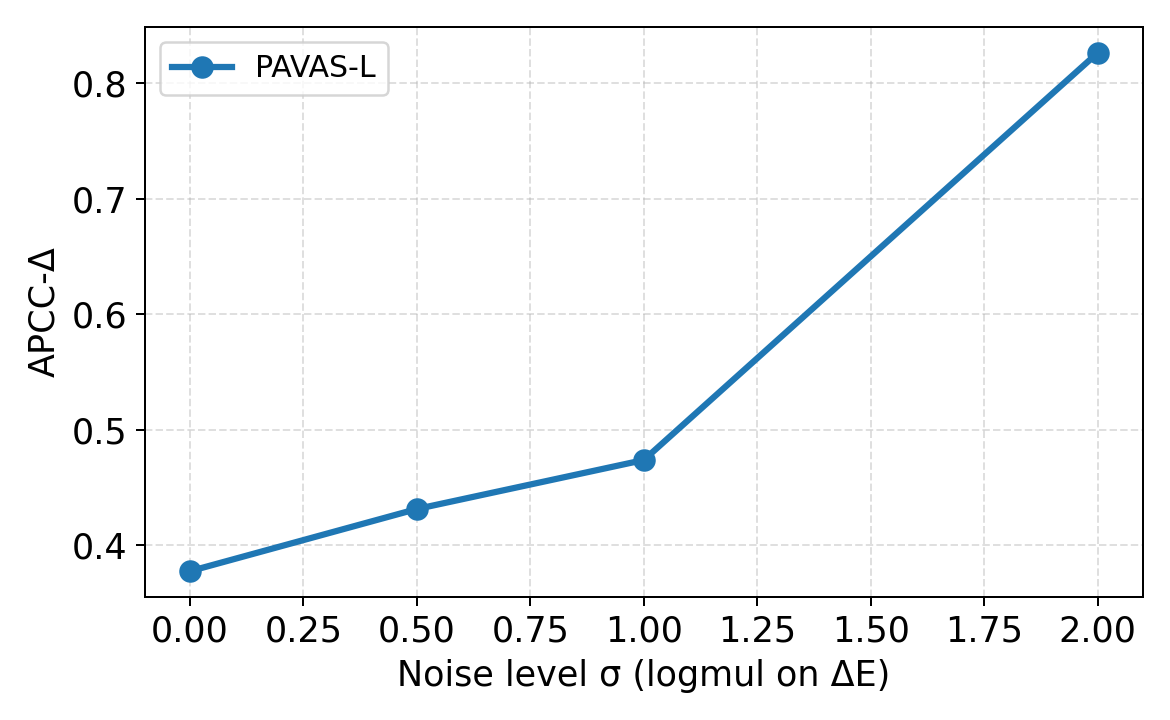}
    \caption{\textbf{APCC-$\Delta$ under injected noise.} As the perturbation magnitude increases, APCC-$\Delta$ degrades progressively, supporting that the metric is sensitive to meaningful changes in the estimated physical quantities.}
    \label{fig:apcc_noise}
\end{figure}

We further analyze whether APCC-$\Delta$ reflects meaningful physics--audio coupling rather than spurious correlation. 
Under inter-class shuffling of the estimated physical magnitude changes, APCC-$\Delta$ becomes substantially worse (0.378 $\rightarrow$ 0.655), indicating that the metric does not arise from chance correlation. 
We additionally inject noise into the estimated physical quantities and observe a progressive degradation in APCC-$\Delta$ as the perturbation magnitude increases (see Fig.~\ref{fig:apcc_noise}). Together, these results support that APCC-$\Delta$ captures meaningful physics--audio coupling and responds consistently to perturbations in the estimated physical quantities.

\section{Module Details}\label{sec:E}
\subsection{Instruction Prompts for VLMs}
\label{supp:vlm_prompts}

We employ a Vision--Language Model (VLM)~\cite{bai2025qwen2} to extract physics-relevant object information through two sequential stages:
(1) \emph{moving-object discovery}, which identifies objects that exhibit genuine physical motion throughout the video, and
(2) \emph{mass estimation}, which predicts a physically plausible mass (in kilograms) for each moving object.
Both stages rely on instruction prompts designed to produce structured, machine-readable JSON output, suitable for downstream processing in the Physics Parameter Estimator (PPE).
Below, we detail the motivation, design principles, and full templates used for each stage.

\paragraph{Prompt for moving-object discovery.}
The first stage requires identifying which entities in the scene undergo meaningful motion over the entire clip.
Because apparent frame-to-frame changes may arise from camera movement or aliasing, the prompt explicitly instructs the VLM to reason at the sequence level and to consider only true object motion driven by translation, rotation, articulation, or deformation.
To guarantee reliable detection, the prompt includes:
(i)~a taxonomy of admissible motion types,
(ii)~explicit exclusion rules for non-solid continuous media (\eg, water, smoke, fire),
and (iii)~strict formatting constraints (plain JSON, no markdown, no invented object names).
The VLM is required to return a JSON object mapping each detected moving object to a short verb phrase describing its action (e.g., \texttt{"basketball":"bouncing"}, \texttt{"door":"opening"}).
A full prompt template is provided in Table~\ref{tab:prompt_moving_objects}.

This structured output plays a dual role: it filters out irrelevant scene elements and provides canonicalized object keys that are passed verbatim to the subsequent mass-estimation stage, ensuring consistency across stages and preventing object-identity drift.

\paragraph{Prompt for mass estimation.}
Given the JSON map of detected moving objects generated by the previous prompt, the second instruction estimates the mass of each object.
The prompt explicitly injects this JSON as the \texttt{MOVING OBJECTS} block and obligates the VLM to:
(i)~use the provided keys without alteration,
(ii)~return a single plain JSON object, and
(iii)~supply, for every object, both a concise rationale and a numeric field \texttt{"weight\_kg"}.

The rationale must be grounded in visual evidence---such as material composition, approximate dimensions, density cues, or human body build---and must end with the same numeric estimate that appears in \texttt{"weight\_kg"}.
The policy section enforces strict structural constraints (no renaming, no additional keys, no markdown, numeric values only) and offers heuristics for estimating mass from typical object specifications.
Several few-shot examples are embedded directly in the prompt to increase output stability and reduce hallucination.
A template is given in Table~\ref{tab:prompt_mass}.

Together, these two prompts enable the VLM to produce consistent, interpretable, and physically grounded object descriptors that serve as input to our Physics Parameter Estimator.
The strict formatting guarantees ensure that the downstream modules can parse and align object identities reliably, allowing robust integration of semantic and physical cues into the overall PAVAS pipeline.

\subsection{Details of Physics Parameter Estimator}
\label{supp:ppe_config}

The Physics Parameter Estimator (PPE) extracts object-level physical attributes from unconstrained visual scenes using three components:
(i) a Vision–Language Model (VLM) for sequence-level motion discovery and object-wise mass estimation,
(ii) an open-vocabulary video segmentation pipeline based on Florence-2~\cite{xiao2024florence} and SAM2~\cite{ravi2024sam}, and
(iii) a CUT3R~\cite{wang2025continuous}-based dynamic 3D reconstruction module for metrically scaled velocity estimation.
All modules operate on the same per-video object keyspace produced by the VLM, ensuring that object descriptions, segmentation tracks, and 3D trajectories remain consistently aligned throughout the pipeline.

\noindent\paragraph{Vision--Language Model.}
Both moving-object discovery and mass estimation are implemented with the \texttt{Qwen/Qwen2.5-VL-7B-Instruct} model\footnote{https://huggingface.co/Qwen/Qwen2.5-VL-7B-Instruct}.  
Input videos are provided through the chat interface as a single video message with adaptive spatial downsampling controlled by a vision-token budget of \texttt{MAX\_PIXELS}=$512{\times}28^2$. In both stages, generation is bounded by \texttt{max\_new\_tokens}=192 and \texttt{min\_new\_tokens}=16.

For moving-object discovery, decoding is performed without stochastic sampling (default greedy decoding) to stabilize key naming and JSON structure. 
For mass estimation, we reuse the same model and pixel scaling but enable stochastic decoding with \texttt{do\_sample=True}, \texttt{temperature}=0.4, and \texttt{top\_p}=0.92 to allow more nuanced and diverse physical reasoning output. 
A truncation heuristic detects incomplete JSON (e.g., unbalanced braces or outputs that terminate near the token limit) and triggers a single regeneration with an expanded budget (\texttt{max\_new\_tokens} multiplied by \texttt{RETRY\_MULTIPLIER=2}). 
In both stages, we parse the model output as a single JSON object.
For moving-object discovery, we further clean the map by lowercasing and discarding clearly invalid entries, whereas for mass estimation we ensure there is exactly one record per input object and convert the \texttt{"weight\_kg} fields into numeric values.

\paragraph{Open-vocabulary video instant segmentation model.}
To obtain open-vocabulary instance masks for each moving object, we couple Florence-2~\cite{xiao2024florence} with the SAM2~\cite{ravi2024sam} video segmentation model. 
For every moving-object key discovered by the VLM~\cite{bai2025qwen2}, we construct a text grounding query in the format used by Florence-2 for open-vocabulary detection (\ie, by prefixing the object name).
Florence-2 applies this query to sparsely sampled keyframes, where the sampling interval is chosen so that the resulting keyframes align with the 8 Frames Per Second (FPS) temporal grid used by the downstream 3D reconstruction pipeline (\ie, CUT3R).

Each detected box is converted into a binary instance mask using the SAM2 image predictor, and these masks initialize the SAM2 video predictor, which leverages XMem-style propagation~\cite{cheng2022xmem} to track object instances, producing temporally consistent mask sequences.
When Florence-2 returns no detection for a keyframe, we insert an empty mask to preserve alignment across time.
The final output is a per-object sequence of instance masks, providing a stable, frame-aligned segmentation track that directly supports subsequent 3D point-cloud fusion and instantaneous velocity estimation.

\paragraph{Dynamic 3D reconstruction model.}
We obtain metrically scaled 3D geometry using CUT3R~\cite{wang2025continuous}, instantiated with the \texttt{cut3r\_512\_dpt\_4\_64.pth} checkpoint\footnote{https://github.com/CUT3R/CUT3R}.
All videos are temporally resampled to 8\,FPS, which aligns the inference of CUT3R with the segmentation sequence.
CUT3R operates on a resolution of \(512\times512\); accordingly, RGB frames are resized to this resolution prior to inference.
The instance masks produced by SAM2 are also resized to the same resolution of CUT3R, which preserves discrete object boundaries while avoiding interpolation artifacts that could distort the mask geometry.

For each frame, CUT3R predicts a dense metric point map together with a per-pixel confidence score.  Points whose confidence is below a fixed threshold ($0.7$) are removed, suppressing geometrically unstable or low-parallax regions while retaining reliable surfaces. The remaining points form a metrically meaningful reconstruction aligned to CUT3R’s estimated camera coordinate system.
Then, object-wise 3D geometry is obtained by indexing the reconstructed point map with each resized instance mask. 
This extracts all 3D points whose originating pixels were assigned to a given object. 
If no valid 3D points remain after confidence filtering, the object is treated as absent for that frame; otherwise, a centroid is computed by averaging the surviving points.

Centroid trajectories are defined over the same 8~FPS temporal grid. 
Instantaneous velocities are computed via finite differences of successive centroids, \ie, Euclidean displacement divided by the frame interval (\(\Delta t = 1/8\)~second).
No temporal smoothing or post-hoc filtering is applied, so the reported velocities directly reflect CUT3R’s raw geometric estimates. 
Frames with missing centroids propagate a \texttt{None} placeholder. 
During downstream conditioning, any velocity that depends on such missing endpoints is replaced with a learnable velocity-occlusion token, ensuring that occluded motion segments are explicitly encoded and do not contaminate the continuous velocity embedding.

\subsection{Handling Missing Temporal Observations}
\label{sec:supp_occlusion_token}
PAVAS uses two learnable occlusion tokens for temporally missing observations. First, in object feature extraction, frames where the object is absent or occluded are replaced with an \emph{object-occlusion token}, yielding a stable object-level conditioning stream across time. Second, in velocity modulation, when a valid velocity cannot be computed for a frame because the required masks or 3D endpoints are unavailable, the corresponding input is replaced with a \emph{velocity-occlusion token}. 
Both occlusion tokens are learnable embeddings with the same dimensionality as their corresponding conditioning streams, allowing missing observations to be handled without changing the conditioning interface.
No mass-occlusion token is introduced, since mass is estimated as a time-invariant scalar for each object rather than a frame-wise quantity. These tokens therefore serve different roles: the object-occlusion token handles missing object observations, whereas the velocity-occlusion token handles unavailable motion estimates.

\subsection{Diffusion Transformer Backbone}
\label{app:sec:multimodal_blocks}

\begin{figure*}
    \centering
    \includegraphics[width=1.0\linewidth]{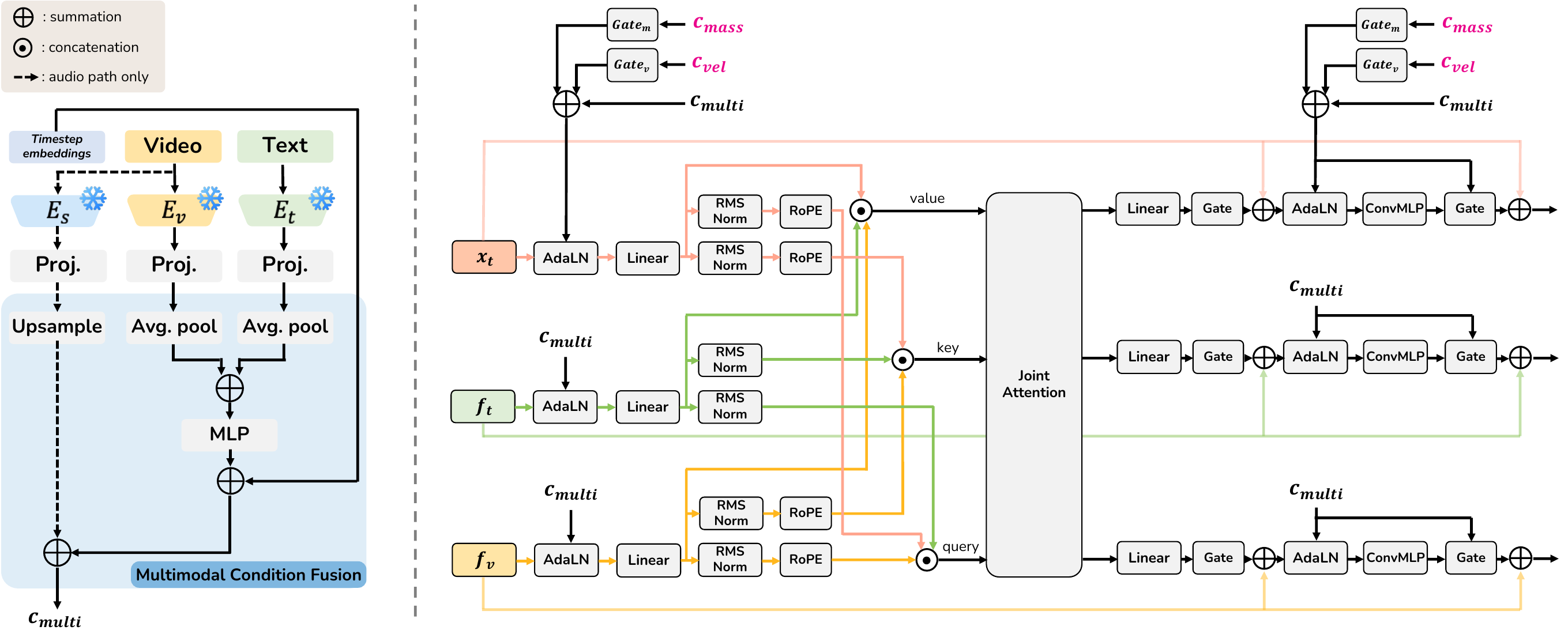}
    \caption{\textbf{Multimodal Diffusion Transformer Blocks and Conditioning Path.}
    \textbf{[Left]} The multimodal condition $\mathbf{c}_{\text{multi}}$ is constructed from synchronization, visual, and textual features together with the diffusion timestep embedding. 
    Note that the upsampled synchronization features are used only when conditioning the audio pathway.
    \textbf{[Right]} Each multimodal diffusion transformer block contains three parallel streams (audio, text, video) that interact through \emph{joint attention}. 
    In contrast, the audio-only (unimodal) blocks used later in the network reuse the same audio pathway but replace joint attention with \emph{self attention}; these blocks are not shown in the diagram but follow directly from the audio branch depicted here. 
    $\mathbf{x}_t$, $\mathbf{f}_t$, and $\mathbf{f}_v$ denote the audio latent, text features, and video features, respectively.
    Physics conditions $\mathbf{c}_{\text{mass}}$ and $\mathbf{c}_{\text{vel}}$ are injected via $\Delta$-modulation, where learnable gates ($\mathbf{Gate}_\mathrm{m}$, $\mathbf{Gate}_\mathrm{v}$) scale the mass and velocity signals and add them residually to the base multimodal condition $\mathbf{c}_{\text{multi}}$, enabling gradual and stable incorporation of physical effects.}
    \label{fig:condition}
\end{figure*}

Our model adopts a diffusion--transformer architecture that integrates audio, vision, text, and synchronization cues within a unified latent space. 
The backbone consists of (i)~a sequence of diffusion transformer blocks that jointly process tokenized audio, visual, and textual inputs, and (ii)~a conditioning pathway that delivers semantic, visual, and physics-aware signals to every block.
Audio is represented as a latent sequence produced by a pretrained Variational Auto Encoder (VAE)~\cite{kingma2013auto}, enabling efficient diffusion while preserving high-fidelity reconstruction.
This section describes the backbone configuration, the formation of the multimodal conditioning stream $\mathbf{c}_{\text{multi}}$, and the physics-specific pathways $\mathbf{c}_{\text{mass}}$ and $\mathbf{c}_{\text{vel}}$. See Fig.~\ref{fig:condition} for a detailed backbone structure.

\paragraph{Audio latent representation.}
Audio modality is modeled in a compact latent domain obtained through a two-stage spectral encoder.
Waveforms are first converted into mel spectrograms via short-time Fourier analysis.
For the 16\,kHz model, we use 80 mel bins with a hop size of 256 samples, yielding a
latent frame rate of 31.25\,FPS.  
For the 44.1\,kHz model, we use 128 mel bins with a hop size of 512 samples,
resulting in a frame rate of 43.07\,FPS.
Each mel frame is then encoded by a pretrained VAE into a 20-dimensional (16\,kHz) or
40-dimensional (44.1\,kHz) latent vector using a 1D convolutional encoder--decoder with a
temporal downsampling factor of~2.  
The VAE adopts magnitude-preserving convolutional and normalization layers, which stabilize
latent statistics without affecting perceptual quality.

The diffusion backbone operates directly on these latent sequences, benefiting from both
reduced temporal length and a smoother optimization landscape.
During inference, generated latents are decoded back into mel spectrograms via the same
VAE and converted to waveforms by a neural vocoder~\cite{lee2022bigvgan}.
These latent representation provide high fidelity while remaining computationally
efficient for diffusion-based generation.

\paragraph{Model variants.}
We train three model capacities that vary in hidden width and depth while sharing the same multimodal block design.
The 16\,kHz model (PAVAS-S) operates on 20-dimensional audio latents with hidden size $h{=}448$ and uses a stack of $(N_M,N_U)=(4,8)$ multimodal and audio-only transformer blocks.
The higher-rate variants (PAVAS-M/L) operate on 40-dimensional latents at 44.1\,kHz and increase the hidden width to $h{=}896$ to accommodate the doubled latent dimension.
The largest model (PAVAS-L) further increases depth to $(N_M,N_U)=(7,14)$, offering the highest generation fidelity.
A summary of architectural differences is provided in Table~\ref{tab:model_variants}.

Table~\ref{tab:pavas_variants} shows that increasing model capacity consistently improves physics correlation and perceptual quality.
PAVAS-M achieves the highest audio quality, while PAVAS-L yields the strongest physics alignment (APCC-$\Delta$) and the rest of perceptual metrics, establishing it as our default model for the main experiment.

\begin{table}[t]
\centering
\resizebox{1.0\linewidth}{!}{
\begin{NiceTabular}{l@{\hspace{10pt}}
                    c@{\hspace{10pt}}
                    c@{\hspace{10pt}}
                    c@{\hspace{10pt}}
                    c@{\hspace{10pt}}
                    c}
\toprule
Model & Sample rate & Latent dim & Hidden dim $h$ & $(N_M, N_U)$ & Params (M) \\
\midrule
PAVAS-S  & 16\,kHz   & 20 & 448 & (4, 8)  & 166 \\
PAVAS-M & 44.1\,kHz & 40 & 896 & (4, 8)  & 630 \\
PAVAS-L & 44.1\,kHz & 40 & 896 & (7, 14) & 1039 \\
\bottomrule
\end{NiceTabular}
}
\caption{\textbf{Summary of model variants.}
Each variant differs in latent dimensionality, hidden width, and the number
of multimodal ($N_M$) and audio-only ($N_U$) transformer blocks.}
\label{tab:model_variants}
\end{table}
\renewcommand{\arraystretch}{1.0}
\begin{table}[t]
\small
\centering
\resizebox{1.0\linewidth}{!}{
\begin{NiceTabular}{l@{\hspace{8pt}}
                    c@{\hspace{8pt}}
                    c@{\hspace{8pt}}
                    c@{\hspace{8pt}}
                    c@{\hspace{8pt}}
                    c}
\toprule
Model 
& {\footnotesize Physics corr.}
& {\footnotesize Distribution match.} 
& {\footnotesize Audio quality} 
& {\footnotesize Semantic align.} 
& {\footnotesize Temporal align.} \\
\cmidrule(lr){2-2}
\cmidrule(lr){3-3}
\cmidrule(lr){4-4}
\cmidrule(lr){5-5}
\cmidrule(lr){6-6}
& 
APCC-$\Delta\downarrow$
& \fdpasst$\downarrow$
& \ispann$\uparrow$
& IB-score$\uparrow$
& DeSync$\downarrow$ \\
\midrule

PAVAS-S 
& 0.412
& 65.67 
& 16.50 
& 29.41 
& 0.448 \\

PAVAS-M 
& 0.395
& 49.08 
& \textbf{17.94} 
& 34.54 
& 0.449 \\

\rowcolor{defaultColor}
PAVAS-L 
& \textbf{0.378}
& \textbf{47.38}
& 17.51 
& \textbf{35.41} 
& \textbf{0.446} \\
\bottomrule
\end{NiceTabular}
}
\caption{\textbf{Quantitative results of PAVAS model variants on VGGSound test split.}
We report physics correlation (APCC-$\Delta$), FD$_\text{PaSST}$, and the perceptual metrics
(audio quality, semantic, and temporal alignment).
PAVAS-L shows the strongest physics alignment and overall most favorable performance across metrics.}
\label{tab:pavas_variants}
\end{table}

\paragraph{Input projection layers.}
Before joint fusion, each modality is projected into the shared hidden dimension:
\begin{itemize}[leftmargin=1.5em]
    \item \emph{Text:} token embeddings are linearly projected and refined by a lightweight MLP.
    \item \emph{Vision:} CLIP patch features are projected and processed by a ConvMLP for local spatial mixing.
    \item \emph{Synchronization cues:} features from the synchronization encoder~\cite{iashin2024synchformer} pass through a 1D convolution, activation, and ConvMLP.
    \item \emph{Audio latents:} VAE-encoded spectrogram latents undergo a temporal convolution and ConvMLP.
\end{itemize}

\paragraph{Temporal layout of synchronization features.}
The synchronization encoder~\cite{iashin2024synchformer} outputs short feature sequences extracted from overlapping video clips.
Since clip-local positions are ambiguous, we introduce a learnable positional embedding per clip window.
After adding this embedding, clip windows are flattened into a single sequence and temporally interpolated to match the audio latent resolution.

\paragraph{Construction of multimodal conditioning $\mathbf{c}_{\text{multi}}$.}
At each diffusion timestep~$t$, the backbone receives a multimodal context vector that summarizes all
non-physical conditioning signals.  
We begin by forming global text and vision descriptors by average-pooling the projected CLIP text
and visual token sequences, and feeding their sum into a MLP:
\[
h_{\text{glob}} = f_{\text{MLP}}\!\big(
    \mathrm{pool}(h_{\text{text}}) + \mathrm{pool}(h_{\text{vision}})
\big).
\]
This global feature is then combined with the diffusion timestep embedding
$f_{\text{time}}(t)$ to produce a sequence-invariant context
$g(t)$, which is broadcast along the temporal dimension.  
The frame-aligned synchronization stream is upsampled to the audio frame rate, yielding
$h_{\text{sync}}(t)$, and the final multimodal conditioning vector is formed as
\[
\mathbf{c}_{\text{multi}}(t) = g(t) + h_{\text{sync}}(t).
\]
Importantly, the upsampled synchronization features $h_{\text{sync}}(t)$ are injected \emph{only}
when conditioning the audio pathway; multimodal blocks use only the $g(t)$ for text and
vision branches.

\paragraph{Multimodal and unimodal blocks.}
The model processes all token streams through $N_M$ multimodal blocks, followed by $N_U$ unimodal (audio path only) blocks.
Each multimodal block performs:
\begin{enumerate}[leftmargin=1.5em]
    \item adaptive normalization of each stream using parameters predicted from
        $\mathbf{c}_{\text{multi}}$, $\mathbf{c}_{\text{mass}}$, and $\mathbf{c}_{\text{vel}}$,
    \item concatenated self-attention over the audio, vision, and text token sequences,
    \item modality-wise splitting of the joint output back into audio, vision, and text streams,
    \item per-stream feed-forward refinement.
\end{enumerate}
Synchronization features influence the block only through $\mathbf{c}_{\text{multi}}$ and do not form a separate token sequence.
Audio-only blocks retain the same conditioning but operate strictly on the audio latent sequence.

\paragraph{$\Delta$-modulation vs.\ Direct summation.}
In the ablation study of Table~3 in the main paper, we consider two strategies of injecting physics conditions into Adaptive Layer Normalization (AdaLN) layer:
\begin{itemize}[leftmargin=1.5em]
    \item \emph{$\Delta$-modulation (ours):}  
    The base AdaLN parameters are predicted from the multimodal condition $\mathbf{c}_{\text{multi}}$. 
    Physics conditions ($\mathbf{c}_{\text{mass}}$ and $\mathbf{c}_{\text{vel}}$) are fed to zero-initialized residual branches that predict only a \emph{correction} to this base modulation. 
    At initialization, these branches produce no effect, and their contribution grows gradually as training learns mass- and motion-dependent adjustments.
    Formally, each block refines its AdaLN parameters as
    $\tilde{\omega} = \omega(\mathbf{c}_{\text{multi}}) + \alpha_m g_m(\mathbf{c}_{\text{mass}}) + \alpha_v g_v(\mathbf{c}_{\text{vel}})$,
    where $g_m,g_v$ are lightweight MLPs and $\alpha_m,\alpha_v$ are learnable gates.
    We show this strategy in the Fig.~\ref{fig:condition}.
    \vspace{1mm}
    \item \emph{Direct summation (ablative baseline):}  
    In this variant, the physics streams are \emph{added directly} to the multimodal condition, without a residual zero-initialized pathway. 
    We form a physics-augmented condition
    $\mathbf{c}_{\text{phys}} = \mathbf{c}_{\text{multi}} + \mathbf{c}_{\text{mass}} + \mathbf{c}_{\text{vel}}$
    and compute modulation parameters as $\tilde{\omega} = \omega(c_{\text{phys}})$.
    This corresponds to a straightforward feature-level mixing of mass and velocity with $\mathbf{c}_{\text{multi}}$, lacking the gradual, gated adaptation provided by $\Delta$-modulation.
\end{itemize}
\vspace{1mm}
By comparing these two modes, we can disentangle the benefit of residual, gate-controlled physics injection from simply augmenting the multimodal condition with additional features.

\subsection{Training Details}
\label{app:training}

\paragraph{Overall procedure.}
PAVAS is trained in a two stage manner.  
The first stage trains a general-purpose multimodal latent diffusion backbone for video-to-audio generation.  
In the second stage, we introduce the Physics Parameter Estimator (PPE) and the Physics-Driven Audio Adapter (Phy-Adapter) and fine-tune the backbone to incorporate mass- and velocity-aware conditioning.

\paragraph{Stage~1: Backbone training.}
Unless stated otherwise, all backbone variants (S/M/L) share the same optimization hyperparameters.  
We use the AdamW optimizer with a learning rate of $1{\times}10^{-4}$, weight decay $1{\times}10^{-6}$, and $(\beta_1,\beta_2)=(0.9,0.95)$.  
A linear warm-up of 1K steps is followed by a cosine decay schedule over 300K iterations, with learning rate reductions at 80\% and 90\% of training, reaching $1{\times}10^{-5}$ and $1{\times}10^{-6}$, respectively.
Mixed-precision (bf16) training and gradient clipping are applied to improve numerical stability.  
We also maintain an exponential moving average (EMA) of the model weights using a post-hoc formulation with a relative width of $\sigma_{\mathrm{rel}} = 0.05$.  
Audio latents and visual/text embeddings are precomputed offline and streamed from disk during training.

\paragraph{Stage~2: Physics-aware fine-tuning.}
During physics integration, the audio, visual, and text encoders remain frozen, while the diffusion transformer blocks, the gating branches, and the physics-conditioning pathways are updated.  
We reuse the same optimizer configuration but shorten training to 30K iterations and reduce the learning rate to $1{\times}10^{-5}$.  
To ensure robustness when motion cues are ambiguous or missing, physics tokens (mass or velocity) are randomly substituted with their corresponding null tokens with probability~0.1.  
This dropout provides the model with implicit classifier-free denoising behavior for the physics pathway.

\paragraph{Batching and compute.}
All models are trained with a global batch size of 512 using distributed data-parallel training on NVIDIA H100 GPUs.  
To keep the compute setting consistent across stages, both stage~1 and stage~2 use the same hardware allocation: PAVAS-S is trained on 2~H100 GPUs, while PAVAS-M and PAVAS-L are trained on 8~H100 GPUs.  
We adopt \texttt{bf16} mixed-precision training to reduce memory footprint and to stabilize optimization at large batch sizes.  
For training efficiency, all audio latents and visual/synchronization features are precomputed offline and streamed from disk during training, avoiding on-the-fly STFT, VAE encoding, or vision forward passes.  
This caching strategy ensures that both 16\,kHz and 44.1\,kHz variants achieve similar throughput despite differences in latent dimensionality and backbone width, and it allows the larger PAVAS-L model to fit comfortably within the memory budget while maintaining stable training dynamics.

\section{Runtime and Invalid/Missing-Observation Statistics of PPE}
\label{sec:F}
We report module-wise runtime profiling and representative statistics of the PPE stack. Runtime is measured on a single H100 GPU after warm-up and averaged over 100 randomly sampled videos from VGGSound. 
Although PAVAS introduces non-trivial computational overhead, it is not intended for real-time operation. 
These results instead characterize the cost of the current pipeline and how often invalid or missing observations arise in unconstrained videos.

\subsection{Module-wise Runtime Profiling}
\label{sec:supp_runtime}
Table~\ref{tab:runtime_profile} reports the average runtime of each component in PAVAS-L. Among the PPE modules, CUT3R~\cite{wang2025continuous} dominates the computational cost, whereas the Phy-Adapter itself adds only a small overhead relative to the backbone. 
This indicates that the main bottleneck lies in extracting physically grounded conditioning signals rather than injecting them into the generator.

\subsection{Invalid and Missing-Observation Statistics}
\label{sec:supp_failure}
We also quantify representative cases of invalid estimates and missing observations in the PPE pipeline. Text--scene mismatches due to VLM hallucination occur in only 0.04\% of samples, and invalid mass outputs occur in 3.55\%. For velocity estimation, CUT3R~\cite{wang2025continuous} produces low-confidence 3D reconstructions in 1.12\% of frames, which are replaced with occlusion tokens. In Grounded-SAM2~\cite{ravi2024sam}, 65.15\% of frames are replaced with occlusion tokens. Importantly, this high ratio mainly reflects natural occlusions and fast motion in in-the-wild videos rather than segmentation failure. Overall, these statistics show that truly invalid estimates are rare, while many missing observations arise from realistic visual ambiguity that PAVAS explicitly handles through fallback tokens (see Sec.~\ref{sec:supp_fallback}).

\begin{table}[t]
\centering
\scriptsize
\resizebox{\linewidth}{!}{
\begin{tabular}{cccccc}
\toprule
VLM (mov.) & VLM (mass) & GSAM2 & CUT3R & Phy-Adapter & Backbone \\
\midrule
0.90 & 0.73 & 0.71 & 4.82 & 0.32 & 1.90 \\
\bottomrule
\end{tabular}
}
\vspace{1mm}
\caption{\textbf{Module-wise runtime of PAVAS-L (seconds per 10-second video).} Runtime is measured on a single H100 GPU after warm-up and averaged over 100 randomly sampled videos from VGGSound. CUT3R~\cite{wang2025continuous} dominates the PPE runtime, while the Phy-Adapter adds relatively small overhead.}
\label{tab:runtime_profile}
\end{table}

\section{Generalization Beyond Impact Scenes}
\label{sec:G}
We additionally compare PAVAS-L with MMAudio-L~\cite{cheng2025mmaudio} on both VGG-NonImpact and VGG-Impact to assess its performance beyond impact-centric scenes. VGG-NonImpact is constructed by excluding the 10 impact-related classes used in VGG-Impact from the VGGSound test split. Table~\ref{tab:nonimpact_vs_impact} shows that PAVAS-L improves over MMAudio-L on most metrics in both settings, suggesting that the benefit of physics-aware conditioning extends beyond impact-heavy events while remaining effective on impact-centric scenes.

\begin{table}[t]
\centering
\small
\resizebox{\linewidth}{!}{
\begin{NiceTabular}{l c c c c}
\toprule
\textbf{Metric} 
& \multicolumn{2}{c}{\textbf{VGG-NonImpact}} 
& \multicolumn{2}{c}{\textbf{VGG-Impact}} \\
\cmidrule(lr){2-3} \cmidrule(lr){4-5}
& \textbf{MMAudio-L} & \textbf{PAVAS-L} 
& \textbf{MMAudio-L} & \textbf{PAVAS-L} \\
\midrule
\fdpasst$\downarrow$      & 62.2 & \textbf{47.9} & 203.4 & \textbf{182.9} \\
\fdpann$\downarrow$       & 4.73 & \textbf{3.92} & 19.8 & \textbf{18.4} \\
\fdvgg$\downarrow$        & \textbf{1.01} & 1.09 & \textbf{4.10} & 5.41 \\
\klpann$\downarrow$       & 1.66 & \textbf{1.56} & 1.67 & \textbf{1.50} \\
\klpasst$\downarrow$      & 1.40 & \textbf{1.36} & 1.29 & \textbf{1.19} \\
\ispann$\uparrow$         & 17.5 & \textbf{17.6} & \textbf{4.72} & 4.45 \\
IB-score$\uparrow$        & 33.1 & \textbf{35.4} & 33.9 & \textbf{36.0} \\
DeSync$\downarrow$        & \textbf{0.44} & 0.45 & \textbf{0.30} & 0.31 \\
\bottomrule
\end{NiceTabular}
}
\vspace{1mm}
\caption{\textbf{Quantitative evaluation on non-impact and impact sets.} PAVAS-L improves over MMAudio-L on most metrics in both VGG-NonImpact and VGG-Impact, suggesting that the benefit of physics-aware conditioning is not limited to impact-centric scenes.}
\label{tab:nonimpact_vs_impact}
\end{table}

\section{Setup of User Study}\label{sec:H}
To complement the objective evaluations, we conduct a user study to assess the perceptual quality of the generated audio. 
We compare PAVAS-L against six state-of-the-art video-to-audio models—See \& Hear~\cite{xing2024seeing}, V-AURA~\cite{viertola2025temporally}, VATT~\cite{liu2024tell}, V2A-Mapper~\cite{wang2024v2a}, TARO~\cite{ton2025taro}, and MMAudio-L~\cite{cheng2025mmaudio}—selected for their strong performance across semantic, temporal, and distributional metrics. 
We sample eight non-speech clips from the VGGSound test set, excluding low-resolution or ambiguous videos. 
For each clip, participants evaluate seven model outputs played back-to-back over the same visual content, with the ordering randomized to avoid bias. 
A total of 27 participants take part in the study, resulting in 1,512 individual ratings.

Participants rate each model output independently using a 5-point Likert~\cite{likert1932technique} scale (1--5; strongly disagree to strongly agree). 
After each audio segment finishes, they pause the video and answer four questions corresponding to the following evaluation criteria:

\begin{itemize}[leftmargin=1.3em]
    \item \textbf{Audio Quality.}
    \emph{``Rate whether the audio sounds clear, natural, and free of distracting noise or artificial artifacts. Ignore the visual content and judge the audio alone.''}

    \item \textbf{Semantic Alignment.}
    \emph{``Rate whether the type of sound matches the events or actions shown in the video (\eg, whether the produced sound is appropriate for the depicted scenario).''}

    \item \textbf{Temporal Alignment.}
    \emph{``Rate whether audio events occur at the correct time relative to visible events (\eg, impacts, collisions, or actions). Misalignment includes delayed, early, or repeated events.''}

    \item \textbf{Physical Plausibility.}
    \emph{``Rate whether the audio reflects the physical properties observed in the video—such as mass, material, speed, and impact strength. Sounds should feel physically consistent with the visual motion and contact forces.''}
\end{itemize}

Table~2 in the main paper summarizes the results. PAVAS-L achieves the highest mean score across all four aspects, with particularly strong gains in the newly introduced physical plausibility dimension.

\section{Additional Discussion on Occlusion and Off-Screen Audio}
\label{sec:I}
PAVAS does not assume that sounding objects remain perfectly visible and trackable throughout the video. In practice, missing object or motion cues often arise due to natural occlusions, fast motion, or unreliable reconstruction in unconstrained videos. 
To handle such cases, the model uses occlusion tokens as fallback inputs rather than requiring complete physical observations at every frame. Moreover, PAVAS can still generate off-screen sound sources through its text-conditioned generation pathway, similar to prior V2A systems that use text prompts (\eg, MMAudio~\cite{cheng2025mmaudio}).
This design allows the model to benefit from explicit physical conditioning when reliable cues are available, while remaining robust when such cues are partially missing.

\section{Additional Qualitative Samples}\label{sec:J}
\begin{figure*}
    \centering
    \includegraphics[width=1.0\linewidth]{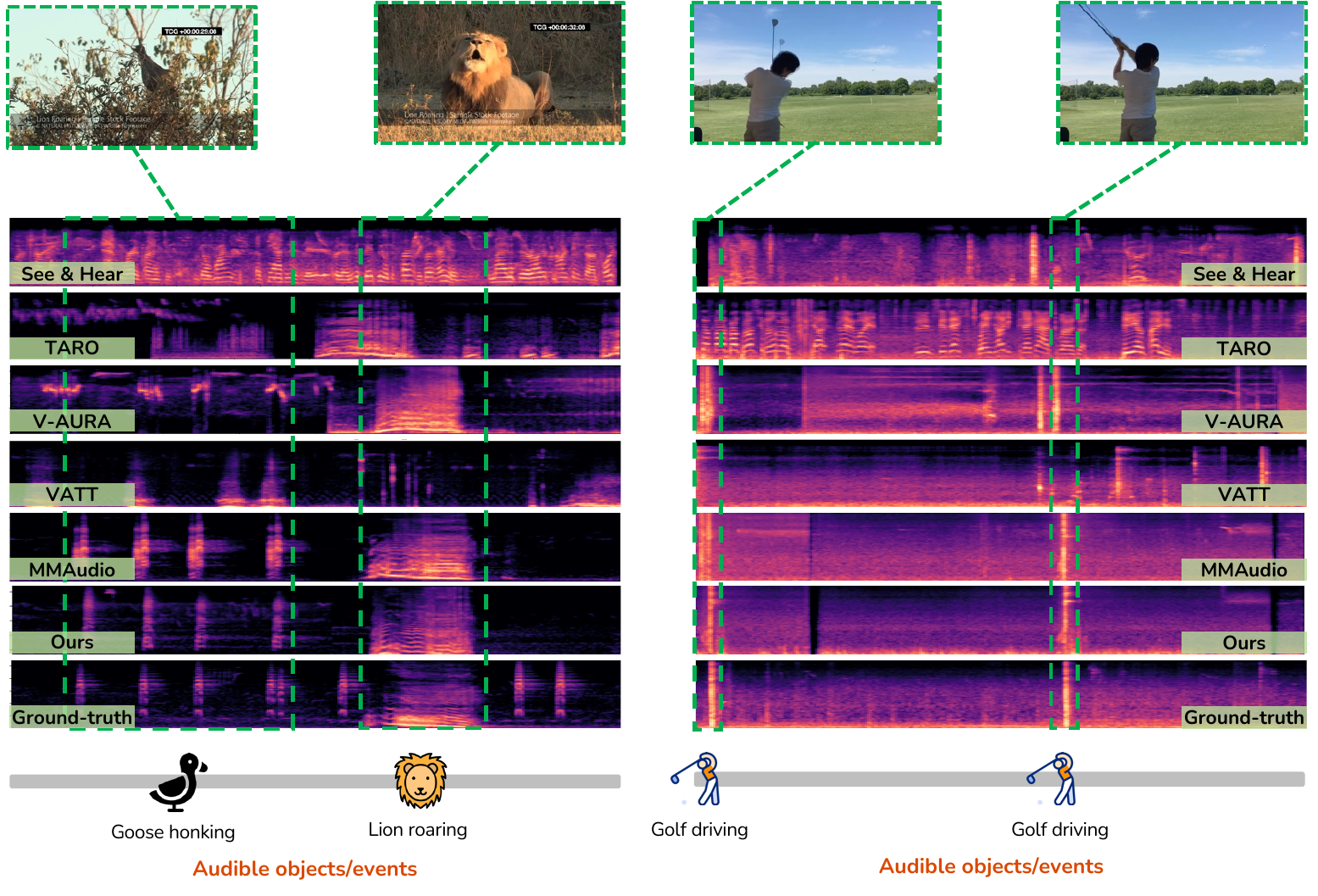}
    \vspace{-1.5mm}
    \caption{
    \textbf{Qualitative comparison of generated spectrograms.}
    We present spectrogram visualizations from the state-of-the-art video-to-audio generation models~\cite{xing2024seeing,ton2025taro,viertola2025temporally,liu2024tell,cheng2025mmaudio}, ours, and the ground-truth audio on VGGSound test split.
    Green dashed lines highlight spectral structures that correspond to visually observable events, and icon markers indicate audible objects or interactions.
    \textbf{[Left]} PAVAS produces spectral patterns that follow the timing and duration of the visual events more faithfully. 
    For example, during the goose honking moment, our model avoids generating unnaturally prolonged honks—a pattern observed in MMAudio~\cite{cheng2025mmaudio}.
    \textbf{[Right]} In the golf driving example, V-AURA~\cite{viertola2025temporally} and MMAudio~\cite{cheng2025mmaudio} incorrectly generate \emph{two} impact sounds at the beginning of the sequence. 
    In contrast, ours produces a single, sharply localized impact that matches the ground-truth spectrogram.
    \vspace{-1mm}}
    \label{fig:supp_qual_(a)}
\end{figure*}

\begin{figure*}
    \centering
    \includegraphics[width=1.0\linewidth]{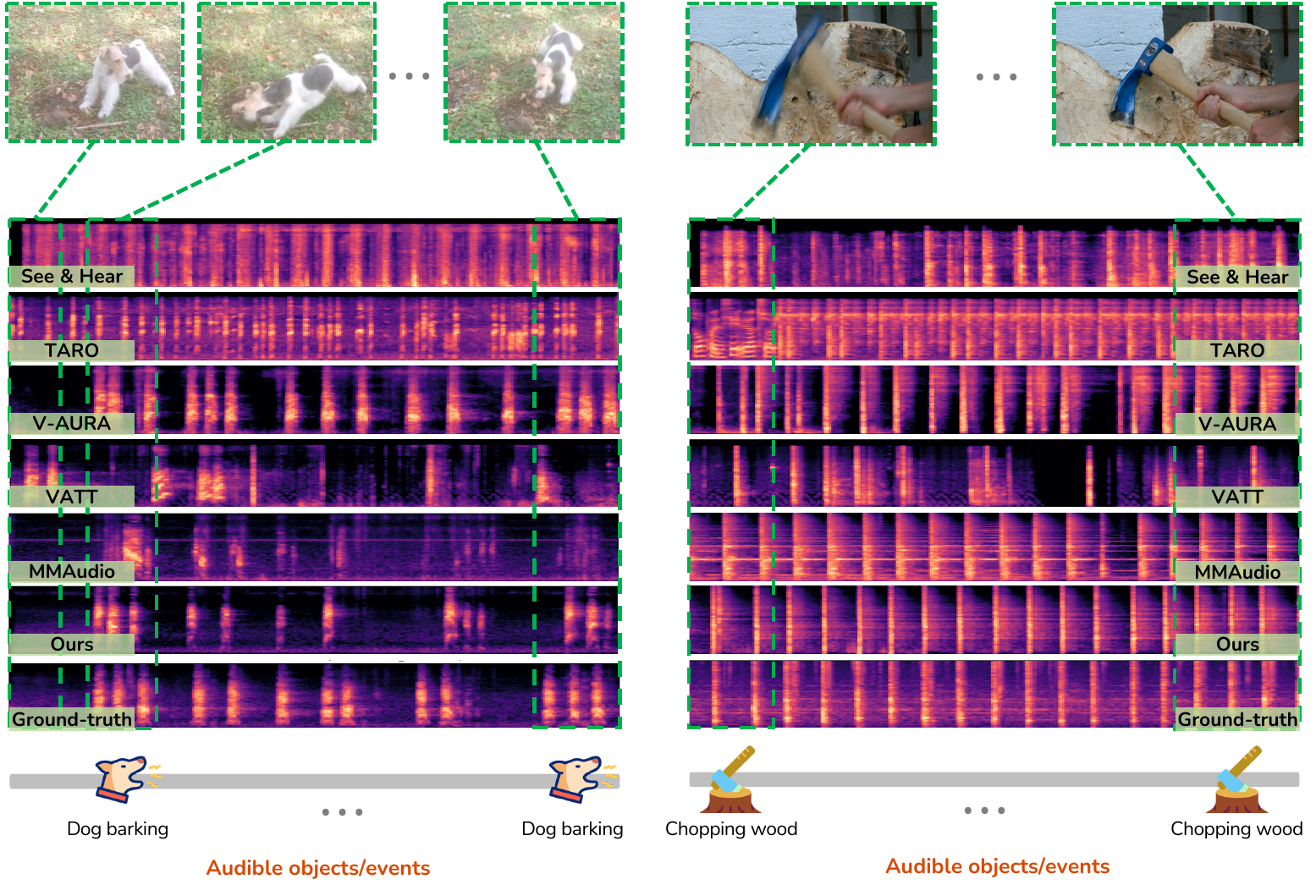}
    \vspace{-1.5mm}
    \caption{
    \textbf{Qualitative comparison of generated spectrograms.}
    We present spectrograms produced by state-of-the-art video-to-audio models~\cite{xing2024seeing,ton2025taro,viertola2025temporally,liu2024tell,cheng2025mmaudio}, PAVAS, and the ground truth. 
    Green dashed lines mark spectral patterns that correspond to visual events, and icon markers denote audible objects or interactions in the scene.
    \textbf{[Left]} In the dog-barking example, PAVAS most accurately reproduces the timing and spectral shape of the bark. 
    Competing models either miss the barking moment or fail to emit a bark at all when the dog opens its mouth.
    \textbf{[Right]} In the wood-chopping scene, PAVAS generates impact transients that are temporally aligned with the chopping motion and exhibit spectral structures closely matching the ground truth.
    Other models tend to produce misaligned strikes or spectrogram patterns that deviate noticeably from the sharp, short-decay broadband signature of real impact sounds.
    }
    \vspace{-1mm}
    \label{fig:supp_qual_(b)}
\end{figure*}

To complement the quantitative results in the main paper, we provide 
additional qualitative comparisons in Figs.~\ref{fig:supp_qual_(a)} and~\ref{fig:supp_qual_(b)}, 
highlighting how PAVAS differs from recent Video-to-Audio (V2A) generation 
models~\cite{xing2024seeing,ton2025taro,viertola2025temporally,liu2024tell,cheng2025mmaudio}. 
For each example, we visualize mel-spectrograms generated by 
state-of-the-art baselines alongside our method and the ground-truth 
audio. We annotate each figure with (i)~green dashed lines to indicate 
spectral structures that correspond to visually observable events, and 
(ii)~icon markers denoting audible objects or interactions present in the 
scene.

Across the qualitative samples, ours better matches the timing, duration, and spectral shape of audio events depicted in the video. 
In the goose honking and golf-driving examples (Fig.~\ref{fig:supp_qual_(a)}), competing 
methods often produce temporally duplicated impacts or unnaturally 
prolonged vocalizations. 
In contrast, ours aligns its transient events 
with the visual dynamics and avoids over-extending acoustic energy when 
the visual cue indicates only a short-lived action.
Figure~\ref{fig:supp_qual_(b)} further demonstrates this. 
In the dog-barking scene, existing V2A models fail to emit a bark at the moment 
when the dog opens its mouth, or they produce temporally shifted 
patterns. 
PAVAS instead generates a bark whose onset and spectral 
structure closely follow the ground truth. 
A similar observation holds in the wood-chopping example: our method produces sharply localized, short-decay broadband transients characteristic of real impacts, whereas other methods generate misaligned strikes or smeared spectral patterns.

Overall, these qualitative examples illustrate how incorporating 
physics-aware cues into the diffusion backbone helps PAVAS generate 
audio that is not only temporally and semantically consistent with the 
video, but also physically plausible in terms of impact strength, 
material response, and motion-dependent timing.

\begin{table*}[t]
\centering
\caption{\textbf{Prompt used for moving-object discovery in the Physics Parameter Estimator (PPE).}}
\label{tab:prompt_moving_objects}
\begin{tabular}{p{16cm}}
\toprule
Return only a plain JSON object (no markdown, no code fences, no extra text) that maps the moving objects in the video to their visible state or action. \\[0.6em]
\\
\textbf{\textit{Definition of a moving object (sequence-level):}} \\
- Determine motion over the entire clip (not frame-by-frame). \\
- Count an object as moving if any of the following holds: \\
\hspace*{1.2em}(a) Translation: position changes relative to the scene/background (parallax-aware). \\
\hspace*{1.2em}(b) Rotation/orientation change: sustained rotation or turning (e.g., door opening, car turning). \\
\hspace*{1.2em}(c) Articulation or deformation driven by the object or its operator (e.g., walking, running, jumping, swinging, throwing, opening, closing, lifting, pushing, pulling, rolling, bouncing, striking, strumming, typing, pedaling). \\
- Do not count as moving: motion caused only by camera pan/zoom/tilt/shake; microscopic flicker or aliasing; motion that is ambiguous or below perceptual threshold. \\[0.6em]
\\
\textbf{\textit{Format:}} \\
- Keys are short lowercase noun phrases naming moving objects; include simple attributes for disambiguation (e.g., \texttt{runner in striped shirt}, \texttt{red car}). \\
- Values are short gerund verb phrases (1--3 words), optionally with a direct object (e.g., \texttt{drifting}, \texttt{pushing cart}, \texttt{opening door}). \\[0.6em]
\\
\textbf{\textit{Objects to exclude as keys:}} \\
- Continuous media: water (ocean, river, waves), atmospheric effects (smoke, fog, clouds), fire, precipitation (rain, snow, hail), wind/storms, volcanic material (lava, ash). \\
- Clothing/accessories as standalone objects (shirts, hats, logos, patterns); these may appear only as attributes to describe an entity. \\[0.6em]
\\
\textbf{\textit{Output format:}} \\
\texttt{\{"object": "state/action", ...\}} \\[0.6em]
\\
\textbf{\textit{Examples:}} \\
\texttt{\{"red car":"drifting","black pickup truck":"colliding"\}} \\
\texttt{\{"basketball":"bouncing","player":"running"\}} \\
\texttt{\{"door":"opening","person":"pushing"\}} \\
\texttt{\{"runner in striped shirt":"sprinting","black dog":"chasing"\}} \\
\texttt{\{"shopping cart":"rolling","man":"pushing"\}} \\

\bottomrule
\end{tabular}
\end{table*}
\begin{table*}[t]
\centering
\caption{\textbf{Prompt used for mass estimation in the Physics Parameter Estimator (PPE).}}
\label{tab:prompt_mass}
\begin{tabular}{p{16cm}}
\toprule
You will be shown a short video. From this video, moving objects and their visible state/action are provided as a JSON map,\quad \texttt{MOVING OBJECTS: } \texttt{\{"object": "state/action", ...\}} \\[0.6em]
\\
\textbf{\textit{Task:}} \\
For every key in \,\texttt{MOVING OBJECTS}, estimate its mass in kilograms and return exactly one JSON object of the form: \\
\texttt{\{"object":\{"rationale":"<concise explanation ending with the same number>","weight\_kg":<number>\}, ...\}} \\[0.8em]
\\
\textbf{\textit{Policy:}} \\
- Use only the keys given in \texttt{MOVING OBJECTS}; do not invent, rename, split, or merge keys. \\
- Include every key exactly once, in the same order as in \texttt{MOVING OBJECTS}. \\
- Keys must match \texttt{MOVING OBJECTS} verbatim, character-for-character. \\
- \texttt{"weight\_kg"} must always be a numeric value (no units, no text). \\
- Rationales should be concise, reference visible attributes (e.g., material, dimensions, body build), and end with the same number reported in \texttt{"weight\_kg"}. \\
- Avoid round anchor numbers unless clearly justified. \\
- Base estimates on visible material, thickness, density, and overall volume in the video. For standardized items use specs (convert lb$\rightarrow$kg by $\div 2.20462$). \\
- Return only the final JSON object. No markdown, no code fences, no extra text. \\[1.0em]
\textbf{\textit{Examples:}} \\[-0.2em]

\texttt{MOVING OBJECTS: \{"black pickup truck":"colliding","red car":"drifting"\}} \\
\texttt{OUTPUT: \{"black pickup truck":\{"rationale":"Mid-size double-cab pickup; steel frame → about 1925.0 kg","weight\_kg":1925.0\},} \\
\texttt{\hspace*{4.0em}"red car":\{"rationale":"Compact sedan; 4-door body → about 1410.0 kg","weight\_kg":1410.0\}\}} \\[0.6em]

\texttt{MOVING OBJECTS: \{"basketball":"bouncing","player":"running"\}} \\
\texttt{OUTPUT: \{"basketball":\{"rationale":"Official size 7 ball; inflated rubber bladder → about 0.62 kg","weight\_kg":0.62\},} \\
\texttt{\hspace*{4.0em}"player":\{"rationale":"Adult male; medium build → about 72.0 kg","weight\_kg":72.0\}\}} \\[0.6em]

\texttt{MOVING OBJECTS: \{"dog":"jumping","suitcase":"rolling"\}} \\
\texttt{OUTPUT: \{"dog":\{"rationale":"Medium-size Labrador; lean build → about 27.0 kg","weight\_kg":27.0\},} \\
\texttt{\hspace*{4.0em}"suitcase":\{"rationale":"Hard-shell carry-on, empty → about 3.8 kg","weight\_kg":3.8\}\}} \\[0.6em]

\texttt{MOVING OBJECTS: \{"guitar":"being strummed","chair":"falling"\}} \\
\texttt{OUTPUT: \{"guitar":\{"rationale":"Full-size acoustic guitar with wooden body → about 2.2 kg","weight\_kg":2.2\},} \\
\texttt{\hspace*{4.0em}"chair":\{"rationale":"Plastic molded chair, lightweight → about 3.5 kg","weight\_kg":3.5\}\}} \\[0.6em]

\texttt{MOVING OBJECTS: \{"quadcopter drone":"hovering","bicycle":"coasting"\}} \\
\texttt{OUTPUT: \{"quadcopter drone":\{"rationale":"Prosumer quadcopter with gimbal; typical takeoff mass → about 0.90 kg","weight\_kg":0.90\},} \\
\texttt{\hspace*{4.0em}"bicycle":\{"rationale":"Aluminum road bike, no racks/fenders → about 9.5 kg","weight\_kg":9.5\}\}} \\

\bottomrule
\end{tabular}
\end{table*}

\end{document}